%% file: main.tex
\documentclass[journal]{IEEEtran}

\usepackage{amsmath,amssymb,amsfonts}
\usepackage{graphicx}
\usepackage{booktabs}
\usepackage{multirow}
\usepackage{rotating}
\usepackage{subcaption}
\usepackage{algorithm}
\usepackage{algpseudocode}
\usepackage{xcolor}
\usepackage{url}
\usepackage{cite}

\graphicspath{{./}}

\newcommand{\Hres}{\mathbf{H}}
\newcommand{\vy}{v_y}
\newcommand{\vx}{v_x}

\begin{document}

\title{Diffusion-Residual Model Predictive Steering Control for Vehicle Stabilization at the Limit of Handling under Model Uncertainty}

\author{Bongsob Song%
  \thanks{This work has been submitted to the IEEE for possible publication.
    Copyright may be transferred without notice, after which this version may no longer be
  accessible.}%
  \thanks{The author is
    with the Department of Mobility Engineering, Ajou University, Suwon, South Korea
  (e-mail: bsong@ajou.ac.kr).}%
  \thanks{Manuscript submitted July 8, 2026.}%
\thanks{The companion code reproducing the paper's main quantitative results will be made publicly available at \protect\url{https://github.com/Vehicle-Intelligence-and-Control-Lab/DresMPC-matlab} upon acceptance.}}

\markboth{Diffusion-Residual Model Predictive Steering Control}%
{Song: Diffusion-Residual Model Predictive Steering Control}

\maketitle

\begin{abstract}
  At the limit of handling, a stabilizing model predictive controller (MPC) depends on the yaw-rate
  reference it tracks and the stable-handling envelope it enforces. Both the achievable reference and the
  safe envelope margin are operating-point-dependent and unknown a priori. Fixed or worst-case settings are
  therefore either too conservative or unsafe. The control objective is directional stabilization---bounding
  the side-slip angle to avoid loss of control---measured by peak side-slip. We learn this uncertainty with a
  conditional diffusion residual model and apply it to the controller's reference and constraints rather than
  its control law. Conditioned on the steering command, the model returns the mean and a predictive spread of
  the nominal-model residual. The mean re-sizes the tracked yaw reference; the spread, propagated over the
  prediction horizon, tightens the stable-handling envelope through a one-sided chance back-off. Together
  these form the proposed diffusion-residual MPC (D-res), so that caution is anticipated ahead of the
  tracking error rather than corrected after it by a high-gain loop. In our evaluation the mean re-size
  carries the bulk of the side-slip reduction and recovers the low-friction divergence, while the
  second-moment back-off adds an anticipatory, self-gating envelope margin and a closed-loop-audited
  (aggregate) risk calibration that a mean-only reference leaves open.
  Because only two moments of the residual are required per command, the generator is tabulated offline and
  the online controller adds a single table lookup to the baseline MPC, running within the $100$\,Hz budget
  on an embedded automotive processor (a measured worst-case $4.08$\,ms per step, $41\%$ of the $10$\,ms
  budget, on an NVIDIA Jetson AGX Xavier), with no in-loop diffusion. Across a seven-degree-of-freedom
  (7-DOF) model and high-fidelity CarMaker co-simulation---an in-domain, simulation-scoped evaluation with
  the generator retrained per tier---spanning vehicle, tire, road, and maneuver diversity, D-res reduces
  peak side-slip where the fixed
  bicycle model is least accurate. It also restores directional stability on low-friction maneuvers, where the fixed reference
  over-commands the available grip.
\end{abstract}


\IEEEpeerreviewmaketitle

\input{sections/01-intro}
\input{sections/02-related}
\input{sections/03-problem}
\input{sections/04-method}
\input{sections/05-experiments}

\input{sections/07-conclusion}

\appendices
\section{Vehicle Fleet and Test Maneuvers}\label{app:setup}
The four-vehicle CarMaker~13 fleet spans the vehicle-size axis, with tire and load variants as listed in
Table~\ref{tab:fleet}. The maneuvers are the standardized
ISO/FMVSS handling tests of Table~\ref{tab:scenarios} (open-loop steering inputs except the driver-tracked
ISO~3888 courses of M1/M2), run on dry road and, for M1/M3/M6, also at
low friction (snow, $\mu\!\approx\!0.3$), the dry/low-$\mu$ split of Tables~\ref{tab:cmp-7dof}--\ref{tab:cmp-cm}.

The seven maneuvers probe complementary handling regimes. M1 (ISO~3888-1) is the standard
obstacle-avoidance double lane change; M2 (ISO~3888-2, the ``moose'' test) runs a tighter, more severe
course at higher lateral demand as an emergency-evasion test. M3 (step steer)
probes the transient yaw response, M4 (steady circular) the understeer limit, M5 (sine-with-dwell) the
FMVSS~126 yaw stability, M6 (frequency sweep) the handling frequency response, and M7 (brake-in-turn) the
combined braking-and-cornering stability.

\begin{table}[!t]
  \centering
  \caption{Diversity-study conditions in CarMaker~13: vehicle \emph{size}, \emph{tire}, and
  \emph{load}.}
  \label{tab:fleet}
  \renewcommand{\arraystretch}{1.2}
  \setlength{\tabcolsep}{5pt}
  \footnotesize
  \begin{tabular}{@{}lllcc@{}}
    \toprule
    Class & Configuration & CarMaker 13 & Mass [kg] & Wheelbase [m] \\
    \midrule
    \multirow{4}{*}{Size}
    & Midsize sedan & BMW~5 & $1600$ & $2.98$ \\
    & Compact car & DemoCar & $1463$ & $2.54$ \\
    & Sports car & Audi~TT  & $1500$ & $2.50$ \\
    & SUV & BMW~X5          & $2275$ & $2.98$ \\
    \midrule
    \multirow{2}{*}{Tire}
    & Tire~A & 225/55R17 & \multicolumn{2}{c}{TireRT, base} \\
    & Tire~B & 195/65R15 & \multicolumn{2}{c}{TireRT, low-$\mu$} \\
    \midrule
    \multirow{2}{*}{Load}
    & 1-pax & 1 passenger &  $70$ & - \\
    & 5-pax & 5 passengers & $350$ & - \\
    \bottomrule
  \end{tabular}

  \vspace{2pt}
  {\scriptsize\raggedright TireRT is one of CarMaker's selectable tire models---a Magic-Formula--based
  real-time formulation; the controller instead predicts with a linear tire (Section~\ref{sec:related}).\par}
\end{table}

\begin{table}[!t]
  \centering
  \caption{Standardized ISO/FMVSS test maneuvers and their quantitative test parameters (entry speed,
  radius, steering,~braking).}
  \label{tab:scenarios}
  \renewcommand{\arraystretch}{1.2}
  \setlength{\tabcolsep}{4pt}
  \scriptsize
  \begin{tabular}{@{}llll@{}}
    \toprule
    ID & Maneuver & Standard & Parameters \\
    \midrule
    M1 & double lane change (DLC) & ISO 3888-1 & $80$\,km/h \\
    M2 & severe DLC (moose)    & ISO 3888-2 & $60$\,km/h \\
    M3 & step steer            & ISO 7401   & $80$\,km/h, $\approx\!4^\circ$ step \\
    M4 & steady circular       & ISO 4138   & $R{=}50$\,m, $18{\to}90$\,km/h ramp \\
    M5 & sine-with-dwell       & FMVSS 126  & $80$\,km/h, $0.7$\,Hz \\
    M6 & frequency sweep       & ISO 7401   & $80$\,km/h, $0.1$--$1.1$\,Hz \\
    M7 & brake-in-turn         & ISO 7975   & $100$\,km/h, $R{=}100$\,m, $0.4$g \\
    \bottomrule
  \end{tabular}
\end{table}

\section*{Acknowledgment}
The author thanks Hyunsu Kim, a graduate student at Ajou University, for his assistance with the on-target
deployment and timing measurements.

\bibliographystyle{IEEEtran}
\bibliography{refs}

\end{document}

%% file: sections/01-intro.tex
\section{Introduction}\label{sec:intro}

\begin{figure}[t]
  \centering
  \IfFileExists{fig_overview.png}{%
  \includegraphics[width=0.88\columnwidth]{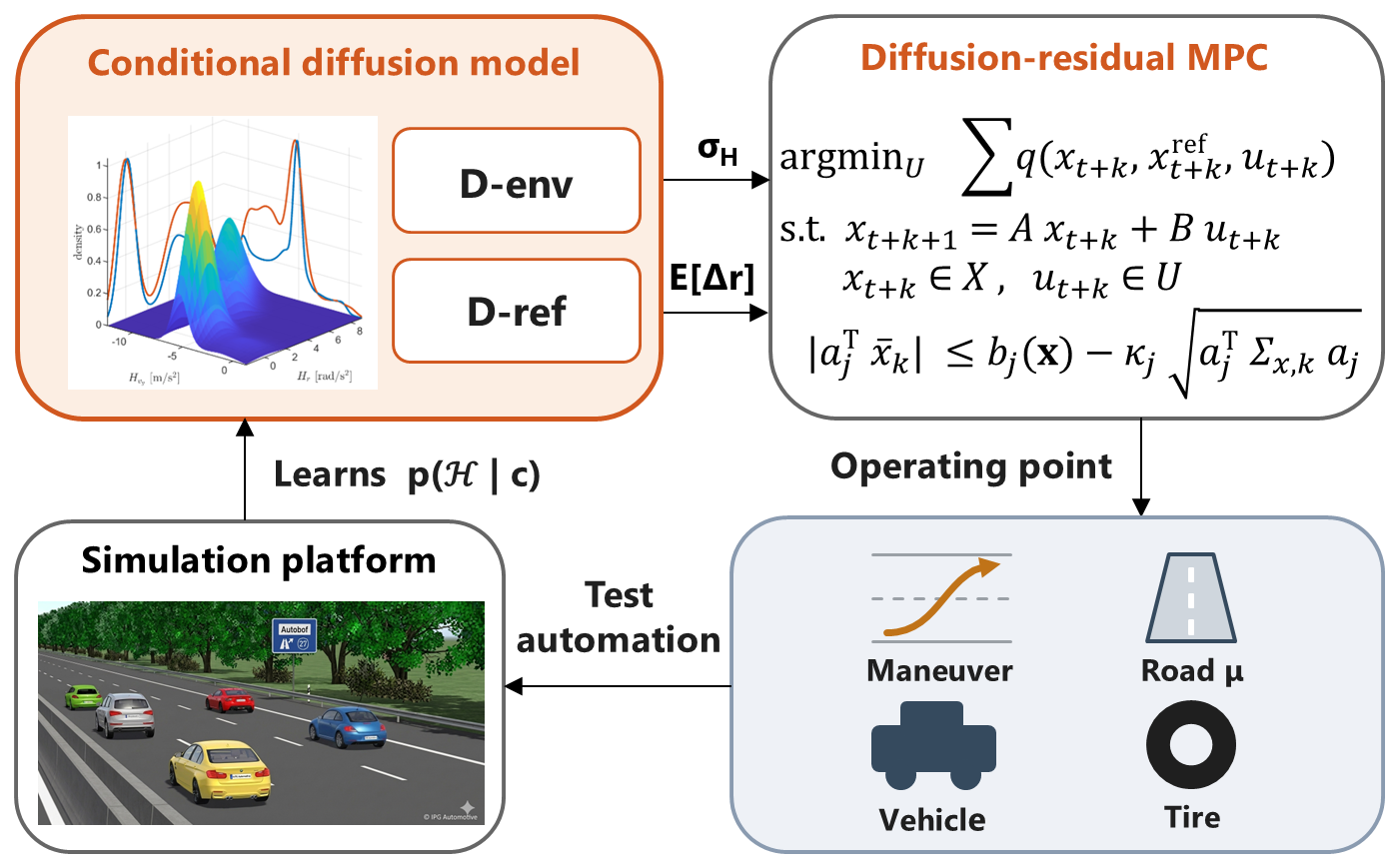}}{%
    \fbox{\parbox[c][0.62\columnwidth][c]{0.98\columnwidth}{\centering\itshape
  [fig\_overview.png pending --- export the \texttt{method\_overview} slide to PNG]}}}
  \caption{Overview of the diffusion-residual MPC and its simulation data pipeline.}
  \label{fig:overview}
\end{figure}

\IEEEPARstart{T}{he} \emph{limit of handling} is the regime in which a road vehicle operates
close to the friction limit of its tires: the lateral-force characteristic saturates, the yaw
dynamics turn nonlinear and become prone to divergence, and the margin for recovery collapses. It is
the regime that decides the outcome of an emergency maneuver---an evasive lane change, a
corner entered at excessive speed on a slippery road---and decades of active-safety engineering have been
aimed at keeping the vehicle controllable there. Two actuator subsystems dominate.
\emph{Active front steering} (AFS) reshapes the yaw response through automatic corrections at the
front axle~\cite{falcone2007predictive}; \emph{electronic stability control} (ESC) restores yaw
stability by braking individual wheels to generate a corrective moment~\cite{vanzanten2000esp}.
Each subsystem commands one actuator
and operates effectively within its own authority, but at the limit the lateral and longitudinal demands
compete for the \emph{same} tire-friction budget, so operating them in isolation forfeits both
performance and stability.

Model predictive control (MPC) has become the dominant framework for stabilization at this
limit, solving a constrained, receding-horizon optimization over a vehicle model that enforces the
actuator limits and a stable-handling envelope explicitly. We work in this predictive,
constraint-based setting and focus on its AFS layer, the controller that
reshapes the yaw response within the shared friction budget. Our objective throughout is directional
stabilization---holding the vehicle within the stable-handling envelope at small side-slip, the
loss-of-control--avoidance regime---as distinct from drift controllers that track a high-sideslip
equilibrium and impose no side-slip
bound~\cite{hindiyeh2014drift,goh2020drift,djeumou2025drift}. An MPC's closed-loop performance, however, is
decided less by the optimizer than by how its objective and constraints are posed---above all the
reference its cost tracks and the margins that define its feasible envelope---and setting these well
is calibration-intensive, motivating dedicated work on MPC
tuning~\cite{garriga2010tuning,forbes2015industry}.

This burden is inseparable from \emph{model uncertainty}: the achievable reference and the safe
envelope margin are both operating-point-dependent and unknown a priori, so with an imperfect model
any \emph{fixed} setting is either too conservative or too aggressive. Robust and stochastic MPC make this
tension explicit, tightening the constraints by an amount dictated by the assumed
uncertainty~\cite{mayne2005robust,mesbah2016stochastic,hewing2020lbmpc}, but at the limit such
worst-case sizing restores guarantees only by shrinking the very envelope that stabilization needs.
We instead address the uncertainty by \emph{learning} it, applying the learned residuals to the
controller's \emph{reference and constraints} rather than its control law. A
diffusion model, conditioned on the speed and steering command, learns two residuals of the nominal model
and returns each as a mean and a predictive spread. The safe
operating region is the Beal--Gerdes stable-handling envelope~\cite{bealgerdes2013mpc}, a state-space
polytope whose facets bound the side-slip, rear-slip, and friction-limited yaw rate. Propagating the
\emph{state-residual} spread over the prediction horizon, we enforce this polytope as a chance constraint,
tightening each facet by a one-sided back-off (\emph{diffusion-envelope MPC}, D-env). The \emph{mean} of the
reference residual re-sizes the tracked reference (\emph{diffusion-reference MPC}, D-ref). Together these
form the proposed \emph{diffusion-residual MPC} (D-res), developed in Section~\ref{sec:method} and
evaluated in Section~\ref{sec:dres}.

The main contributions of this paper, summarized in Fig.~\ref{fig:overview}, are as follows. \emph{(i)}~We
propose D-res (= D-env + D-ref), a learned chance-constrained stabilizing MPC that applies the
conditional-diffusion residual to the controller's constraints and reference rather than its control law.
\emph{(ii)}~The Gaussian-moment constraint tightening and variance propagation follow the standard
cautious-MPC form~\cite{hewing2020cautious,mesbah2016stochastic}, and a calibrated uncertainty has been
propagated into chance constraints before~\cite{vaskov2024frictionadaptive}; the distinction here is the
\emph{source}---a single command-conditioned diffusion generator supplying \emph{both} moments to one
controller---and to our knowledge this is the first use of a diffusion model's predictive spread as a
chance-constraint back-off for limit-handling stabilization. \emph{(iii)}~The controller needs only two
moments of the residual as a function of the command, so the generator is tabulated offline and adds only a
table lookup---a ${\approx}37$\,kB data footprint---to the baseline
MPC, with no in-loop diffusion; we \emph{demonstrate} on-target real-time execution within the $100$\,Hz
budget on an automotive-grade embedded processor (NVIDIA Jetson AGX Xavier, Section~\ref{sub:realtime}).
\emph{(iv)}~We validate it across the vehicle$\times$tire$\times$road$\times$maneuver
diversity of Tables~\ref{tab:fleet} and~\ref{tab:scenarios} in Appendix~\ref{app:setup}---coverage prior
limit-handling studies address only piecewise (Table~\ref{tab:priorwork})---on a
7-DOF model~\cite{rajamani2012vehicle} and CarMaker~13~\cite{carmaker} over ISO/FMVSS maneuvers, delineating
where the learned residual transfers and where it does not. Throughout, each claim is scoped to its
evidence: the evaluation is in-domain and simulation-based, fleet results are medians with per-cell
exceptions disclosed, and the diffusion advantage over a Gaussian baseline concentrates on the
high-fidelity tier.

The remainder of the paper is organized as follows. Section~\ref{sec:related} reviews related work and the
nominal model, Section~\ref{sec:problem} states the problem, Section~\ref{sec:method} develops D-res,
Section~\ref{sec:dres} evaluates and ablates it, and Section~\ref{sec:conclusion} concludes.

%% file: sections/02-related.tex
\section{Related Work}\label{sec:related}

\subsection{Nominal model and uncertainty}\label{sub:estimation}
The controller of this work rests on the standard 2-DOF bicycle
model~\cite{rajamani2012vehicle}; the nominal model with uncertainty is written as
\begin{equation}\label{eq:bicycle}
  \dot x = \big(A+\Delta A\big)\,x + \big(B+\Delta B\big)\,u,
\end{equation}
where $x=[v_y\ r]^\top \in \mathbb{R}^2$ are the state variables with lateral velocity $v_y$ and yaw rate $r$,
and $u=\delta \in \mathbb{R}$ is the control input, the front-wheel steer $\delta$.
The nominal single-track matrices are
\begin{subequations}\label{eq:bicyclemats}
  \begin{align}
    A(v_x)&=
    \begin{bmatrix}
      -\dfrac{C_f+C_r}{m v_x} & -v_x-\dfrac{C_f\ell_f-C_r\ell_r}{m v_x}\\[1.6ex]
      -\dfrac{C_f\ell_f-C_r\ell_r}{I_z v_x} & -\dfrac{C_f\ell_f^2+C_r\ell_r^2}{I_z v_x}
    \end{bmatrix},\label{eq:bicyclemats-A}\\[1ex]
    B&=
    \begin{bmatrix}\dfrac{C_f}{m} & \dfrac{C_f\ell_f}{I_z}
    \end{bmatrix}^\top,\label{eq:bicyclemats-B}
  \end{align}
\end{subequations}
where $v_x$ is the longitudinal speed, $m$ is the vehicle mass, $I_z$ is the yaw moment of
inertia, $\ell_f,\ell_r$ are the distances from the center of gravity to the front/rear axle,
and $C_f,C_r$ are the front/rear axle cornering stiffnesses.

The matrices~\eqref{eq:bicyclemats} are nominal; the true vehicle adds the unknown,
operating-point-dependent perturbations $\Delta A,\Delta B$ of~\eqref{eq:bicycle}, whose size and
shape change along four axes that a validation must therefore span. The \emph{vehicle} sets the mass,
inertia, and load transfer that the two in-plane DOF omit, so the effective stiffnesses vary across
vehicles; the \emph{tire} fixes the force--slip curve, of which $F_y=C\alpha$ is only the
small-slip tangent of a nonlinear, friction-scaled characteristic~\cite{pacejka2012tire} that no
constant $(C_f,C_r)$ tracks; the \emph{road} enters solely through the friction coefficient
$\mu$---never explicit in $A,B$---so identical steer and slip map to markedly different forces as $\mu$
changes; and the \emph{maneuver} selects which of these the controller meets,
steady cornering probing the understeer limit and sine-with-dwell the transient stability boundary.
Because the stiffnesses and the $\mu$ they implicitly
encode enter only a few entries ($C_r$ in $A$, $C_f$ in both $A$ and $B$), the perturbations act
through exactly those terms, and $\mu$ alone has driven extensive work on friction-adaptive
control~\cite{vaskov2024frictionadaptive}.

How a controller accounts for $(\Delta A,\Delta B)$ spans three strategies.
\emph{Re-identification} approaches keep the model current by adapting its parameters to the
operating point, through linear parameter-varying (LPV)/polytopic scheduling~\cite{li2021polytopic} or adaptive,
set-membership parameter updates~\cite{lorenzen2019robust}. The nominal linear time-varying (LTV) MPC this work builds on
already incorporates the simplest such rule in the loop: a \emph{deterministic friction-circle}
re-identification that rescales both cornering stiffnesses by the measured friction
usage~\cite{rajamani2012vehicle,pacejka2012tire},
\begin{equation}\label{eq:nmpc-stiff}
  C_{i}^{\mathrm{eff}}=C_{i}\,k_d,\qquad
  k_d=\sqrt{\max\!\big(k_{\min}^2,\ 1-\rho^2\big)},
\end{equation}
where $i{=}f,r$, $\rho=\min\!\big(\bar\rho,\ \sqrt{a_y^2+a_x^2}/(\mu g)\big)$ is the friction usage from the
measured accelerations $(a_y,a_x)$, bounded at $\bar\rho{=}0.98$, and $k_{\min}{=}0.45$ floors the effective
stiffness---a single lumped factor applied equally to both axles---so it does not collapse as the tires
saturate. This one in-loop adaptation is deployed \emph{identically} by every controller compared here, so
any reported difference isolates the learned residual rather than this rescaling. Like the LPV and adaptive schemes, the lumped scalar still lags the abrupt growth of the mismatch
toward saturation and leaves load transfer uncaptured, which motivates the learned residual below.

\emph{Bounding} approaches, by contrast, confine the mismatch to a set or an a priori distribution
and optimize against the worst case---robust and tube MPC~\cite{mayne2005robust,bemporad1999robust},
the scenario approach~\cite{calafiore2006scenario}, and stochastic or chance-constrained
MPC~\cite{carvalho2015stochastic,calafiore2006distributionally}. Because
$(\Delta A,\Delta B)$ are operating-point dependent rather than adversarial, however, a set wide enough to
cover load transfer, tire saturation, and the full friction range shrinks exactly the
$(\beta,r)$ envelope that limit stabilization needs.

\emph{Learning} the residual keeps the interpretable nominal model and corrects only its residual
error: Gaussian-process (GP) MPC propagates a learned posterior into cautious
constraints~\cite{hewing2020cautious,scampicchio2025gpmpc}, iterative and learning MPC refine the
model across repetitions~\cite{rosolia2018lmpc}, and neural or diffusion
residuals augment a nominal model under a stability-tolerant feedback
law~\cite{shi2019neural,dronediffusion2025,djeumou2025drift}. These consume
the residual as a \emph{point} estimate, however, correcting the model without a calibrated,
operating-point-conditioned measure of how far it can then be trusted. The closest exception is
friction-adaptive stochastic nonlinear MPC (NMPC)~\cite{vaskov2024frictionadaptive}, which \emph{does} propagate a
calibrated uncertainty into chance constraints---but of a Bayesian friction estimate that enters through
the tire--road friction parameters rather than the operating-point residual as a whole. This work instead learns the residuals that
$(\Delta A,\Delta B)$ induce---a state-derivative residual whose calibrated spread sizes the chance
back-off, and a reference residual $\Delta r^{\mathrm{ref}}$ whose mean re-sizes the tracked yaw
reference---as operating-point-conditioned \emph{distributions} (Section~\ref{sec:method}). It
thereby captures tire saturation, load transfer, and road friction \emph{jointly} rather than a single
friction parameter. Realizing it with a diffusion model, and the contrast with generated-plan and
world-model couplings, is taken up in Section~\ref{sub:diffusion}.

\subsection{Predictive steering control at the limit of handling}
Predictive steering control is the dominant framework for stabilization at the limit. Linear
time-varying MPC for active front steering established receding-horizon stabilization
for autonomous vehicles~\cite{falcone2007predictive}, and coordinated
steering-and-braking MPC posed directly in the tire-slip domain extended it to combined
actuation~\cite{dicairano2013yaw}. Beal and Gerdes formalized control to a
\emph{stable-handling envelope}, keeping the vehicle inside a region of the $(\beta,r)$
state space where the tires retain controllability~\cite{bealgerdes2013mpc}.

The nominal controller our study builds on unifies these threads as a velocity-form,
condensed LTV-MPC on the speed-scheduled model~\eqref{eq:bicycle}: a yaw-rate-reference
stabilization objective in the direct-yaw-control tradition~\cite{rajamani2012vehicle,dicairano2013yaw,vanzanten2000esp},
realized through active-front-steering actuation~\cite{falcone2007predictive} and constrained to a
Beal--Gerdes stable-handling envelope~\cite{bealgerdes2013mpc}. At~each~step~it~solves
\begin{subequations}\label{eq:nominal-mpc}
  \begin{align}
    \min_{\{\Delta\delta_k\},\,\xi\ge0}\ &
    \sum_{k=1}^{N_p} Q\,(r_k-r^{\mathrm{ref}}_k)^2
    +\!\sum_{k=0}^{N_c-1}\!\big(R\,\Delta\delta_k^2+S_\delta\,\delta_k^2\big)+w_\xi\xi^2
    \label{eq:nmpc-cost}\\
    \text{s.t.}\ \ & x_{k+1}=A_d(\vx)\,x_k+B_d\,u_k,
    \label{eq:bicycle-d}\\
    & |\delta_k|\le\delta_{\max},\quad |\Delta\delta_k|\le\Delta\delta_{\max},
    \label{eq:nmpc-steer}\\
    & |\beta_k|\le\beta_{\max}+\xi,
    \label{eq:nmpc-beta}\\
    & \Big|\tfrac{\ell_r}{\vx}r_k-\beta_k\Big|\le\alpha_{r,\max}+\xi,
    \label{eq:nmpc-rslip}\\
    & |r_k|\le \tfrac{\mu g}{\vx}+\xi.
    \label{eq:nmpc-yaw}
  \end{align}
\end{subequations}
The cost~\eqref{eq:nmpc-cost}, summed over a prediction horizon $N_p$ and control horizon $N_c$, tracks the
yaw rate to its reference (weight $Q$) while penalizing
steering rate and effort ($R,S_\delta$) and the envelope slack ($w_\xi$); its velocity-form
parameterization follows the active-front-steering MPC of~\cite{falcone2007predictive}. The yaw
reference $r^{\mathrm{ref}}$ is not measured but synthesized from the steady-state
understeer-gradient relation of the same single-track model~\cite{rajamani2012vehicle},
\begin{equation}\label{eq:rref}
  r^{\mathrm{ref}}:=r^{\mathrm{lin}}=\frac{\vx\,\delta}{L+K_{us}\vx^2},\quad
  K_{us}=\frac{m}{L}\Big(\frac{\ell_r}{C_f}-\frac{\ell_f}{C_r}\Big),
\end{equation}
with wheelbase $L=\ell_f+\ell_r$ and understeer gradient $K_{us}$, saturated to the
friction-circle limit $|r^{\mathrm{ref}}|\le\mu g/\vx$. This linear
feedforward is the same understeer relation used to steer at the limit in autonomous
racing~\cite{kritayakirana2012limits}; it holds only while tire force grows with slip and
over-predicts the achievable yaw once the tires saturate. Because $K_{us}$ and that limit depend
on $(m,C_f,C_r,\ell_f,\ell_r,\mu)$, a fixed, mis-identified model mis-sizes the target
across the vehicle/tire/road diversity of Section~\ref{sec:problem}.

The prediction dynamics~\eqref{eq:bicycle-d} are the zero-order-hold discretization of~\eqref{eq:bicycle}
scheduled on $\vx$~\cite{rajamani2012vehicle}, with both axle stiffnesses friction-adapted by the
deterministic friction-circle factor~\eqref{eq:nmpc-stiff}. The input constraints~\eqref{eq:nmpc-steer}
impose the steering magnitude and rate limits of~\cite{falcone2007predictive}.

The Beal--Gerdes stable-handling envelope~\cite{bealgerdes2013mpc,bobier2013staying} is formed by
two slip-state boundaries: the rear-slip constraint~\eqref{eq:nmpc-rslip} keeps the rear tire
unsaturated, and the friction-limited yaw constraint~\eqref{eq:nmpc-yaw} bounds the yaw rate at the
friction-circle cornering limit---the \emph{yaw-acceleration nullcline} of the phase-plane
envelope~\cite{bobier2013staying}. These bound the body side-slip only \emph{implicitly}: the
canonical envelope includes no independent side-slip facet, limiting $\beta$ through the rear-slip line
and the yaw bound together~\cite{bealgerdes2013mpc,bobier2013staying}. We add an explicit side-slip
box~\eqref{eq:nmpc-beta} ($\pm6^\circ$ on $\beta=\arctan(v_y/v_x)\approx v_y/v_x$) as a conservative
guard that proves redundant under closed-loop control (Section~\ref{sub:cmp-cm}). Bounding
$\beta$ at all reflects a \emph{stabilization} intent rather than drift tracking
(Section~\ref{sec:intro}), the distinction captured by the maneuver column of Table~\ref{tab:priorwork}. The yaw bound~\eqref{eq:nmpc-yaw} is the
$\mu$-dependent bound that the linear reference~\eqref{eq:rref} also targets and where model and road
uncertainty concentrate---hence where our chance back-off acts (Section~\ref{sec:method}). All three
constraints share the slack $\xi$, which keeps the quadratic program (QP) feasible at every step through an exact
soft-constraint penalty~\cite{kerrigan2000soft}---\emph{persistent} QP solvability, not recursive
feasibility of the hard envelope.

This division of labor also fixes the scope of our study and our choice of nominal controller. At
the limit the controllable objective is \emph{directional stabilization}: hold the vehicle on the
driver's intended curvature while keeping the slip state controllable. The yaw rate is therefore the
tracked output---it conveys that curvature through the friction-limited reference~\eqref{eq:rref} and is what
a single front-steer input most directly commands---while the side-slip is instead bounded, since
tracking a second target would over-determine the single
actuator~\cite{bealgerdes2013mpc}. We adopt this controller \emph{unchanged} as the nominal baseline (NOM)---the
canonical stabilization design for this regime, a line later extended to emergency avoidance and
shared control~\cite{funke2017,erlien2016}. Since the residual augmentation of Section~\ref{sec:method} is independent
of this choice, a standard, well-understood NOM isolates its effect rather than confounding it with a
re-engineered~controller.

Among representative limit-handling \emph{stabilization} MPCs---those regulating the vehicle to a
yaw-rate reference or a $(\beta,r)$ handling envelope under model and road uncertainty---this work is
placed by the operating diversity and prediction model on which they were validated
(Table~\ref{tab:priorwork}). Progress-maximizing racing MPC~\cite{liniger2015racing} and sampling-based
information-theoretic MPC~\cite{williams2018mppi} pursue path progress or lap time rather than envelope
stabilization under uncertainty, and so fall outside this axis. Among the stabilization methods, each fixes the vehicle and tire and varies at
most the road on a single maneuver, and only the recent diffusion-drift controller generalizes a
learned \emph{dynamics} model across vehicles and tires~\cite{djeumou2025drift}; none spans the
joint vehicle$\times$tire$\times$road$\times$maneuver diversity of
Table~\ref{tab:priorwork}. This work (bottom row) is the
first to span it, and Section~\ref{sec:problem} makes that mismatch concrete.

\begin{table}[t]
  \centering
  \caption{Operating diversity and control model of representative limit-handling
  stabilization MPCs (refer to Tables~\ref{tab:fleet} and~\ref{tab:scenarios} in
  Appendix~\ref{app:setup} for more detail).}
  \label{tab:priorwork}
  \renewcommand{\arraystretch}{1.15}
  \setlength{\tabcolsep}{3pt}
  \scriptsize
  \begin{tabular}{@{}llccll@{}}
    \toprule
    Method & Model & Veh. & Tire & Road/$\mu$ & Maneuver \\
    \midrule
    AFS-MPC~\cite{falcone2007predictive}            & bicycle       & 1 & 1\,(Pacejka) & low-$\mu$ (ice)  & M1-low \\
    AFS$+$brake~\cite{dicairano2013yaw}             & bicycle       & 1 & 1\,(brush)   & dry, snow        & M1,\,M3 \\
    Envelope~\cite{bealgerdes2013mpc}               & bicycle       & 1 & 1\,(affine)  & low-$\mu$        & M6-low \\
    Stochastic~\cite{carvalho2015stochastic}        & bicycle       & 1 & 1\,(linear)  & ---              & avoid$\dagger$ \\
    Frict.-adapt.~\cite{vaskov2024frictionadaptive} & bicycle       & 1 & 2\,(Pacejka) & dry,\,snow       & M1-high/low \\
    GP-ref~\cite{hewing2020cautious}                & bicy.$+$GP    & 1 & 1\,(Pacejka) & dry              & race$\dagger$ \\
    Diff.\ drift~\cite{djeumou2025drift}            & diffusion     & 2 & $\ge2$       & dry,\,wet        & drift$\dagger$ \\
    \textbf{This work}$^{\ddagger}$ & \textbf{bicy.$+$diff.} & \textbf{4} & \textbf{2\,(Pacejka)} & \textbf{dry/low-$\mu$} & \textbf{M1--M7} \\
    \bottomrule
  \end{tabular}

  \vspace{2pt}
  {\scriptsize\raggedright $\dagger$:\ maneuver outside Table~\ref{tab:scenarios}
  (avoid, race, drift); Beal--Gerdes slalom\,$\approx$\,M6 (low-$\mu$).
  $\ddagger$:\ additionally deployed on-target (Section~\ref{sub:realtime}).\par}
\end{table}

\subsection{Diffusion models for control}\label{sub:diffusion}
Denoising diffusion probabilistic models~\cite{ho2020ddpm} and their score-based
stochastic differential equation (SDE) formulation~\cite{song2021score} have become strong conditional generators and have
recently entered control through several couplings. One learns \emph{dynamics or world
models} and plans through them---diffusion predictive control and its constrained
variants~\cite{dmpc2024,dpcc2024}. A second generates a \emph{control target}---a plan,
action, or reference that a downstream tracker follows---as in diffusion
planners~\cite{janner2022diffuser,ajay2023decisiondiffuser}, the learning analogue of
the classical reference/command governor that reshapes a setpoint to honor
constraints~\cite{garone2017reference}. A third learns a \emph{residual} of a nominal
model and hands it to a controller: DroneDiffusion adds a diffusion residual as a
feedforward term to an adaptive sliding-mode law whose stability holds for any bounded
learning error, so the learned model tightens the tracking accuracy bound rather than the
stability guarantee~\cite{dronediffusion2025}. In the
vehicle domain specifically, Djeumou et al.~\cite{djeumou2025drift} condition a
physics-informed diffusion model on online measurements and embed it in a real-time MPC
to drift several cars across tires and surfaces---the closest prior use of diffusion for
limit handling. Their generator serves as the \emph{prediction model}---per-cycle samples
of tire/dynamics parameters the MPC propagates to track a high-sideslip drift
equilibrium---so the sampled dispersion remains prediction-model variability; no stability
envelope or chance level is enforced. The delta here is \emph{where the second moment
lands}: as a calibrated one-sided back-off on an enforced stable-handling envelope.

These couplings trade off differently. A generated reference or action binds the learned
object to one controller and provides no calibrated uncertainty for a safety envelope; a
residual used as additive feedforward overlaps the robust feedback that already
rejects it, so a worse estimate is absorbed by higher gain---amplifying noise and
chattering the actuator---rather than propagated into the
controller's constraints; and a learned dynamics model for drift targets tracking, not a
stochastic constraint. We instead learn the residual \emph{distributions} the mismatch induces and use
their moments in one controller---a state-derivative residual whose propagated spread backs off the
stable-handling envelope, and a reference residual whose mean re-sizes the yaw reference---so caution scales
with the predicted uncertainty rather than a fixed gain. This second-moment coupling---rather than the
point-estimate residual of the schemes above---is what the diffusion generator enables
(Section~\ref{sec:method}).

%% file: sections/03-problem.tex
\section{Problem Statement}\label{sec:problem}
As summarized in Table~\ref{tab:priorwork}, the more operating diversity a limit-handling controller
must span, the larger the uncertainty of the fixed nominal model~\eqref{eq:bicycle} inevitably becomes.
This section defines the problem along three axes---the \emph{driver} that the correction augments, the
\emph{vehicle} across which it must generalize, and the \emph{road and maneuver} at which the residual grows
fastest---and thereby delimits where the nominal predictive controller~\eqref{eq:nominal-mpc} holds and
where it fails.

\begin{figure}[t]
  \centering
  \includegraphics[width=0.86\columnwidth]{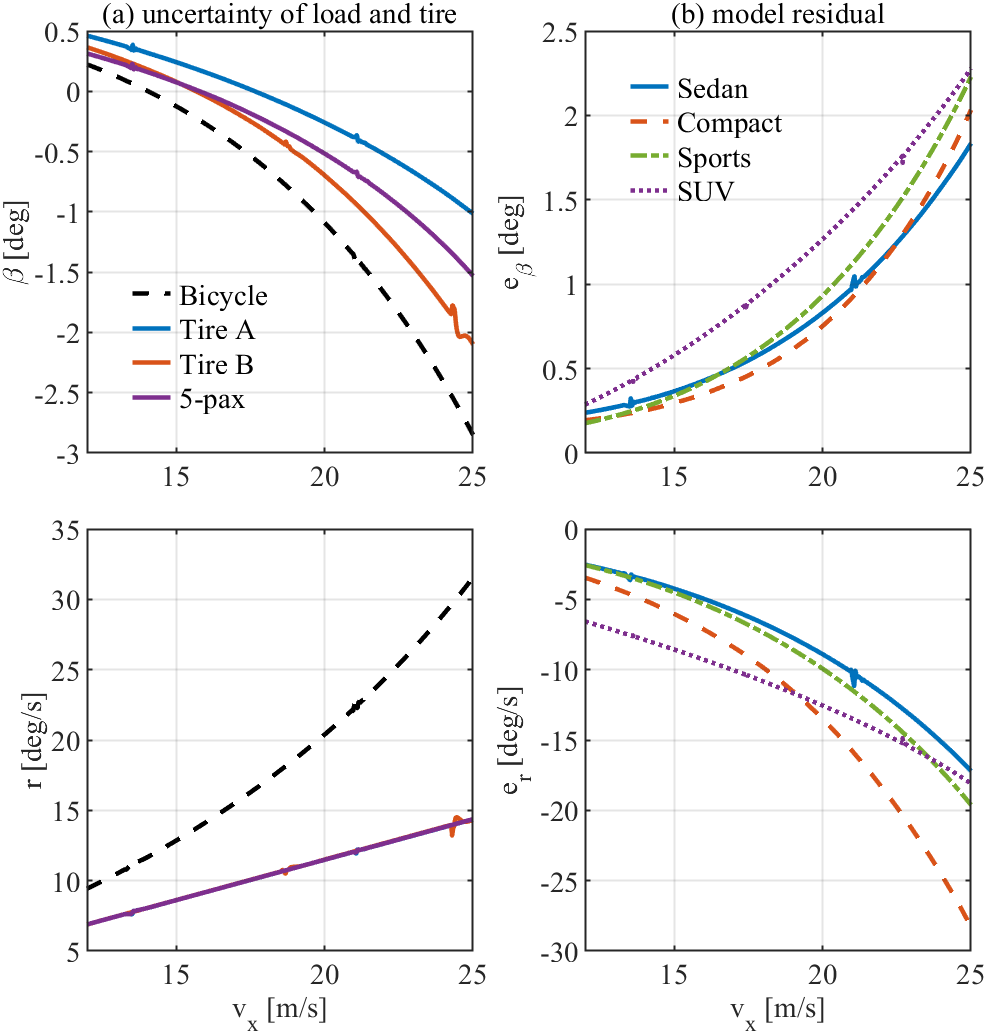}
  \caption{Side-slip ($\beta$), yaw rate ($r$), and residual, in steady cornering. \textbf{(a)}~load and tire: the bicycle model (dashed) against CarMaker (Tire~A, Tire~B, five-occupant load).
  \textbf{(b)}~open-loop state residual $e_\beta,e_r$ across the four-vehicle fleet (Sedan, Compact, Sports, SUV).}
  \label{fig:diversity}
\end{figure}

\subsection{Active front steering with a driver model}\label{sub:afs}
The control authority in this work is AFS: on top of the driver's
steering $\delta_d$ the controller superimposes an auxiliary front road-wheel angle $\delta_c$, so the
front-wheel steer is $\delta=\delta_d+\delta_c$~\cite{falcone2007predictive}. No braking or differential-torque actuation is
used (steering-only); yaw is stabilized through the correction $\delta_c$, while the driver retains path
authority and, through $\delta_d$, sets the tracked yaw reference $r^{\mathrm{ref}}$ in~\eqref{eq:rref}.

Due to this additive structure, AFS performance may depend on how the driver is modeled, so two driver
models are considered. The first is a white-box model
for the 7-DOF tier: a closed-loop Stanley path-tracking
driver, $\delta_d=\theta_e+\arctan\!\big(k\,e_{\mathrm{lat}}/\vx\big)$ with cross-track error
$e_{\mathrm{lat}}$, heading error $\theta_e$, and gain $k{=}1.5$~\cite{thrun2006stanley}.
The second one is a black-box model for the high-fidelity CarMaker tier: the built-in IPGDriver---a
knowledge-based driver---controls longitudinal speed and lateral path-following along each
route~\cite{carmaker}. The \emph{same} predictive
controller in~\eqref{eq:nominal-mpc} generates the correction $\delta_c$ in both tiers; how well this single
formulation transfers across the two distinct driver models is examined in the cross-tier simulation
results. This is the first axis of the problem---the \emph{driver}: a single correction must hold across
two very different driver models.

\subsection{Model uncertainty due to vehicle dynamics}\label{sub:vehicle}
The predictive
controller in~\eqref{eq:nominal-mpc} commands the total front-wheel steer $\delta_k$---the increment
$\Delta\delta_k$ it optimizes, accumulated as
$\delta_k=\delta_{k-1}+\Delta\delta_k$---so the AFS correction applied on top of the driver is
$\delta_{c,k}=\delta_k-\delta_{d,k}$. It is computed on the nominal single-track
model in~\eqref{eq:bicycle}, whose system parameters $(m,I_z,C_f,C_r)$ leave the
operating-point-dependent residual across vehicle, tire, road, and maneuver conditions.
It is shown in Fig.~\ref{fig:diversity}(a) that both side-slip and yaw rate vary with the operating point, {\it i.e.},
the bicycle model in~\eqref{eq:bicycle} against CarMaker models with two tires and
an added load. With the force--slip law $F_y=C\alpha$ valid only at small slip, no constant
$(C_f,C_r)$ tracks the saturating tire. As a consequence, the yaw-rate discrepancy is most pronounced
in Fig.~\ref{fig:diversity}(a): at the limit ($\vx{=}25$\,m/s) the linear bicycle~\eqref{eq:bicycle}
(dashed) over-predicts $r$ at $31^\circ$/s, while all three CarMaker runs---saturating---hold it near
$14^\circ$/s. The side-slip instead \emph{shifts} with the tire and the load: against the bicycle's
$-2.9^\circ$, the base Tire~A holds $\beta$ at $-1.0^\circ$, the lower-friction Tire~B increases it to
$-2.1^\circ$, and a five-occupant ($+280$\,kg) load to $-1.5^\circ$. The residual is thus largest in
$r$ and tire-/load-dependent in $\beta$---a shift the learned correction absorbs in closed loop.

Across the four-vehicle fleet in Table~\ref{tab:fleet} the discarded load transfer and
per-wheel redistribution make the residual class-dependent. The open-loop \emph{state} residual $e_\beta=\beta-\beta^{\mathrm{bic}}$,
$e_r=r-r^{\mathrm{bic}}$ with respect to the bicycle model in the same steady cornering is
plotted in Fig.~\ref{fig:diversity}(b): small at low speed, it grows toward the limit. The yaw-rate residual is largest on the compact
car (peak $|e_r|\!\approx\!28^\circ$/s, $|e_\beta|\!\approx\!2^\circ$), with the sports car and SUV near
$|e_r|\!\approx\!18$--$20^\circ$/s and $|e_\beta|\!\approx\!2.2$--$2.3^\circ$. This is the second
axis---the \emph{vehicle}: its class, tire, and load all shift the
effective stiffnesses, so no fixed correction fits.

\subsection{Limits of MPC under road and maneuver}
Because the road friction $\mu$ never appears in~\eqref{eq:bicycle} but is absorbed into $C_f,C_r$, a
correction identified on one surface is wrong on another. The plant is thus a nonlinear,
parameter-varying map whose mismatch is a \emph{structured} signal whose shape shifts with the operating
point. This is the third axis---the \emph{road and maneuver}: the residual must be \emph{generated
conditioned} on that point rather than fixed in advance.

\begin{figure}[t]
  \centering
  \IfFileExists{fig_traj_featured.png}{%
  \includegraphics[width=0.72\columnwidth]{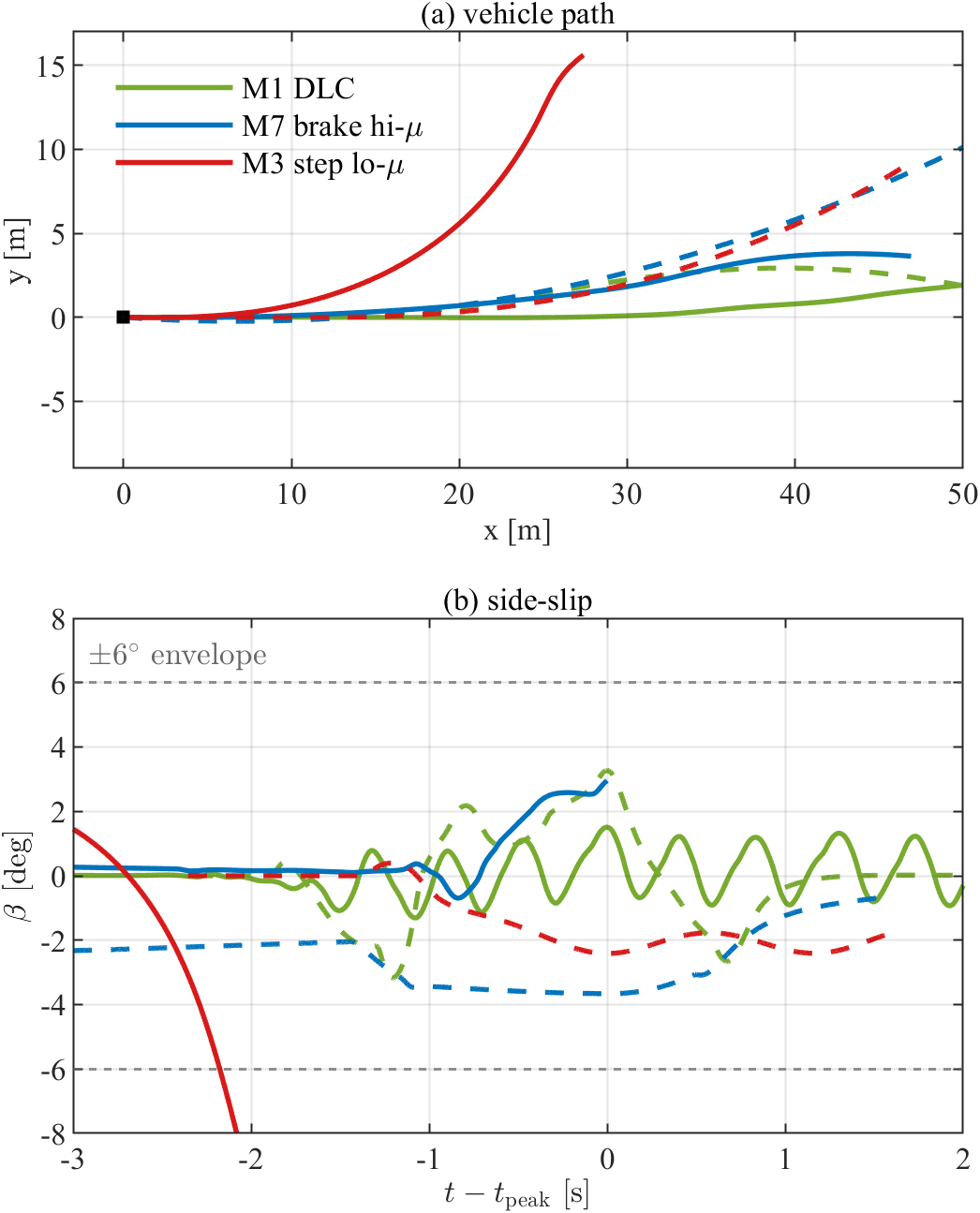}}{%
    \fbox{\parbox[c][0.9\columnwidth][c]{0.98\columnwidth}{\centering\itshape
  [fig\_traj\_featured.png pending]}}}
  \caption{The three featured limit maneuvers (M1, M7, M3-low) under the nominal MPC (design sedan) on both
    evaluation tiers (\emph{solid}: CarMaker, \emph{dashed}: 7-DOF). \textbf{(a)}~Vehicle path in the ground
    plane. \textbf{(b)}~Side-slip $\beta(t)$ against the $\pm6^\circ$ envelope, with time referenced to each
  maneuver's NOM side-slip peak $t_{\mathrm{peak}}$.}
  \label{fig:traj-featured}
\end{figure}

For instance, Fig.~\ref{fig:traj-featured} shows the three featured maneuvers under the nominal MPC on both tiers.
On the mild lane change (M1, refer also to Table~\ref{tab:scenarios}) the CarMaker $\beta$ stays well inside the $\pm6^\circ$ envelope
($\beta_{\mathrm{peak}}\!\approx\!1.5^\circ$, against $3.3^\circ$ on the 7-DOF). On the high-$\mu$ brake-in-turn (M7) it stays inside only
\emph{barely} ($\beta_{\mathrm{peak}}\!\approx\!3.0^\circ$ CarMaker, $3.7^\circ$ 7-DOF), the rear-slip facet~\eqref{eq:nmpc-rslip} nearly
binding. On the low-$\mu$
step (M3-low), commanding against the nominal $\mu{=}1$, the CarMaker controller \emph{diverges}
($\beta_{\mathrm{peak}}\!\approx\!99^\circ$). The tiers agree on the envelope outcome for M1 and M7 but split on the low-$\mu$ step: only the
CarMaker vehicle departs the intended path, while the 7-DOF plant, under-predicting tire saturation, holds
near $2^\circ$ (the cross-tier gap visible in Fig.~\ref{fig:traj-featured}). The failure is largest where the model
error is large and fast---the envelope the nominal MPC enforces is itself built on the linear bicycle, so
its margin degrades exactly where the model does, and at the low-$\mu$ limit the yaw diverges before a
reactive correction can take hold. Above all, this cross-tier gap is what makes the high-fidelity
CarMaker tier indispensable:
the license-free 7-DOF model under-predicts tire saturation and would miss the low-$\mu$ failure on its
own, so only the high-fidelity co-simulation exposes the limit the controller must ultimately survive.

The standard remedy---robust or tube MPC sized for the worst case---restores margin only by shrinking
the usable envelope, forfeiting the limit performance we seek. The remaining limit is one of
\emph{timing}: the error stays near zero and then diverges sharply, so a reactive correction lags the
yaw build-up, whereas a predictive controller committed to a \emph{future} input sequence can apply a
correction generated at the current operating point \emph{before} the error surfaces. The correction
must therefore also be \emph{anticipated ahead} of the error, not merely reacted to.

Given onboard signals $\{v_x,r,a_y,\delta\}$, the nominal model~\eqref{eq:bicycle}, and the controller's
planned input sequence, we seek a real-time steering policy that holds the vehicle inside the
stable-handling envelope at the limit across the tire, mass, and vehicle diversity. It is driven by a
residual \emph{generated conditioned on the operating point}: its \emph{mean} re-sizes the yaw reference
$r^{\mathrm{ref}}$, and its calibrated \emph{spread} sizes a chance back-off on the envelope, so the
controller is cautious exactly where the model is uncertain.

%% file: sections/04-method.tex
\section{Diffusion-Residual MPC for steering control}\label{sec:method}

The proposed controller, D-res, augments the nominal LTV-MPC~\eqref{eq:nominal-mpc} with a shared
conditional-diffusion generator that learns the nominal model's residuals and returns, for each query,
their \emph{mean} and \emph{spread}. In contrast to residual schemes that modify the control law, D-res
applies these moments to the controller's \emph{reference and constraints}. The imposed caution thus
scales with the operating-point uncertainty and is anticipated \emph{before} the tracking error arises
rather than rejected after it. The moments feed the controller's two heads: a \emph{spread},
propagated over the committed prediction horizon, tightens the stable-handling
envelope~\eqref{eq:nmpc-beta}--\eqref{eq:nmpc-yaw} through a chance back-off (D-env), while
the \emph{mean} re-sizes the tracked yaw reference (D-ref); both draw on the \emph{same} generator,
defined next.

\subsection{The shared diffusion residual generator}\label{sub:residual}
Both heads of the proposed controller read \emph{one} conditional diffusion generator that learns
\emph{two heterogeneous residuals} of the nominal bicycle model, each conditioned on the command
$c=[\vx,\delta]$. Throughout, the conditioning input $\delta$ is the \emph{driver's} road-wheel
command---the exogenous yaw demand of Section~\ref{sub:afs}---never the optimized total steer: the MPC's
decision variable does not enter the lookup, so no algebraic loop arises, and the moments are evaluated at
the current measured command, adding no delay. The residual consumed by D-env is the two-channel \emph{state residual} on the
single-track state $x=[\vy\ r]^\top\in\mathbb{R}^2$, the state-derivative mismatch between the bicycle
model and the higher-fidelity measurement,
\begin{equation}\label{eq:residual}
  \Hres=[H_{\vy},H_r]^\top=\dot x^{\mathrm{m}}-\dot x^{\mathrm{nom}},
\end{equation}
where $\dot v_y^m$ follows from the kinematic identity $a_y\!=\!\dot v_y+\vx r$ (the lateral velocity
$\vy$ is unmeasured; estimating it online is a mature literature of its
own~\cite{chindamo2018sideslip}) and $\dot r^m$ from finite-differencing the measured yaw rate,
\begin{subequations}\label{eq:residual-deriv}
  \begin{align}
    \dot x^{\mathrm{m}}_k&=
    \begin{bmatrix}a_{y,k}-v_{x,k}\,r_k\\[3pt]\big(r_{k+1}-r_k\big)/T_s
    \end{bmatrix},\label{eq:rd-meas}\\[4pt]
    \dot x^{\mathrm{nom}}_k&=A(\vx)\,x_k+B\,\delta_k,\label{eq:rd-nom}
  \end{align}
\end{subequations}
with $T_s$ the sample time. Equivalently, this residual is the state-derivative image of the model
perturbation $(\Delta A,\Delta B)$ of~\eqref{eq:bicycle}: for the true plant
$\dot x=(A{+}\Delta A)x+(B{+}\Delta B)\delta$, $\Hres=\dot x^{\mathrm{m}}-\dot x^{\mathrm{nom}}=\Delta A\,x+\Delta B\,\delta$.

A conditional diffusion model represents each residual as a conditional distribution
$p_\psi(\Hres\mid c)$. A single denoiser architecture serves both residuals: a fully-connected network
$\epsilon_\theta(\tilde x_t,t,c)$ with two $\tanh$ hidden layers ($64$ units each) and a linear noise output, fed the
concatenation of the noisy residual $\tilde x_t$, a four-term sinusoidal embedding of the diffusion step
$t$, and the conditioning $c$. Unlike image diffusion's convolutional U-Nets~\cite{ronneberger2015unet},
each target here is a one- or two-dimensional residual with no spatial extent, so a small multilayer
perceptron suffices. Training follows the denoising diffusion probabilistic model (DDPM)
framework~\cite{ho2020ddpm}: a forward process noises a clean sample $\tilde x_0{=}\Hres$ as
$\tilde x_t=\sqrt{\bar\alpha_t}\,\tilde x_0+\sqrt{1-\bar\alpha_t}\,\epsilon$, and $\epsilon_\theta$ is
trained to predict the noise,
\begin{equation}\label{eq:ddpm}
  \mathcal{L}=\mathbb{E}_{t,\epsilon,\tilde x_0}\big\lVert\epsilon-\epsilon_\theta(\tilde x_t,t,c)\big\rVert^2 ,
\end{equation}
which is denoising score matching~\cite{vincent2011connection}. Samples then follow
the DDPM ancestral chain~\cite{ho2020ddpm} from $\tilde x_T\!\sim\!\mathcal N(0,I)$,
\begin{equation}\label{eq:sample}
  \begin{split}
    \tilde x_{t-1}={}&\frac{1}{\sqrt{\alpha_t}}\Big(\tilde x_t-\frac{\beta_t}{\sqrt{1-\bar\alpha_t}}\,
    \epsilon_\theta(\tilde x_t,t,c)\Big)\\
    &+\sqrt{\beta_t}\,z,\qquad z\sim\mathcal N(0,I)\ (t{>}1),
  \end{split}
\end{equation}
over $t{=}T,\dots,1$ ($T{=}50$, a linear schedule $\beta_t\!\in\![10^{-4},\,2\times10^{-2}]$,
$\alpha_t{=}1{-}\beta_t$, $\bar\alpha_t{=}\prod_{s\le t}\alpha_s$; the diffusion-schedule $\beta_t,\alpha_t$
are distinct from the vehicle side-slip $\beta$ and slip angle $\alpha$), each
chain drawing a sample $\Hres^{(s)}\sim p_\psi(\Hres\mid c)$. Learning the score rather than a fixed
parametric form lets the generator capture the residual's \emph{full} conditional structure---its
non-Gaussian shape and operating-point-dependent, heteroscedastic spread---of which the deployed back-off
consumes the per-channel standard deviations~\eqref{eq:sigmaH}.

The generator learns the full distribution, but the stochastic-MPC (SMPC) chance constraints (Section~\ref{sub:mpc})
require only its low-order moments: under a Gaussian moment approximation each boundary chance constraint
tightens in closed form from the predicted mean and covariance. Because the pooled residual is measurably
non-Gaussian (Section~\ref{sec:dres}), this Gaussian form is a data-driven margin-tightening
\emph{surrogate} whose realized risk is audited in closed loop, not an exact per-step probabilistic
guarantee. The moment the envelope back-off (D-env) consumes is the state-residual spread,
\begin{equation}\label{eq:sigmaH}
  \sigma_H=\operatorname{std}[\Hres\mid c]=[\sigma_{H_{\vy}},\sigma_{H_r}],
\end{equation}
estimated from the $S{=}64$ reverse-process draws $\{\Hres^{(s)}\}$ as the square root of the diagonal
of their sample covariance, $\sigma_H=\sqrt{\operatorname{diag}\widehat\Sigma}$ with
$\widehat\Sigma=\tfrac1S\sum_s(\Hres^{(s)}-\bar{\Hres})(\Hres^{(s)}-\bar{\Hres})^\top$.

\begin{figure}[!t]
  \centering
  \IfFileExists{fig_dres_bc.png}{%
  \includegraphics[width=0.72\columnwidth]{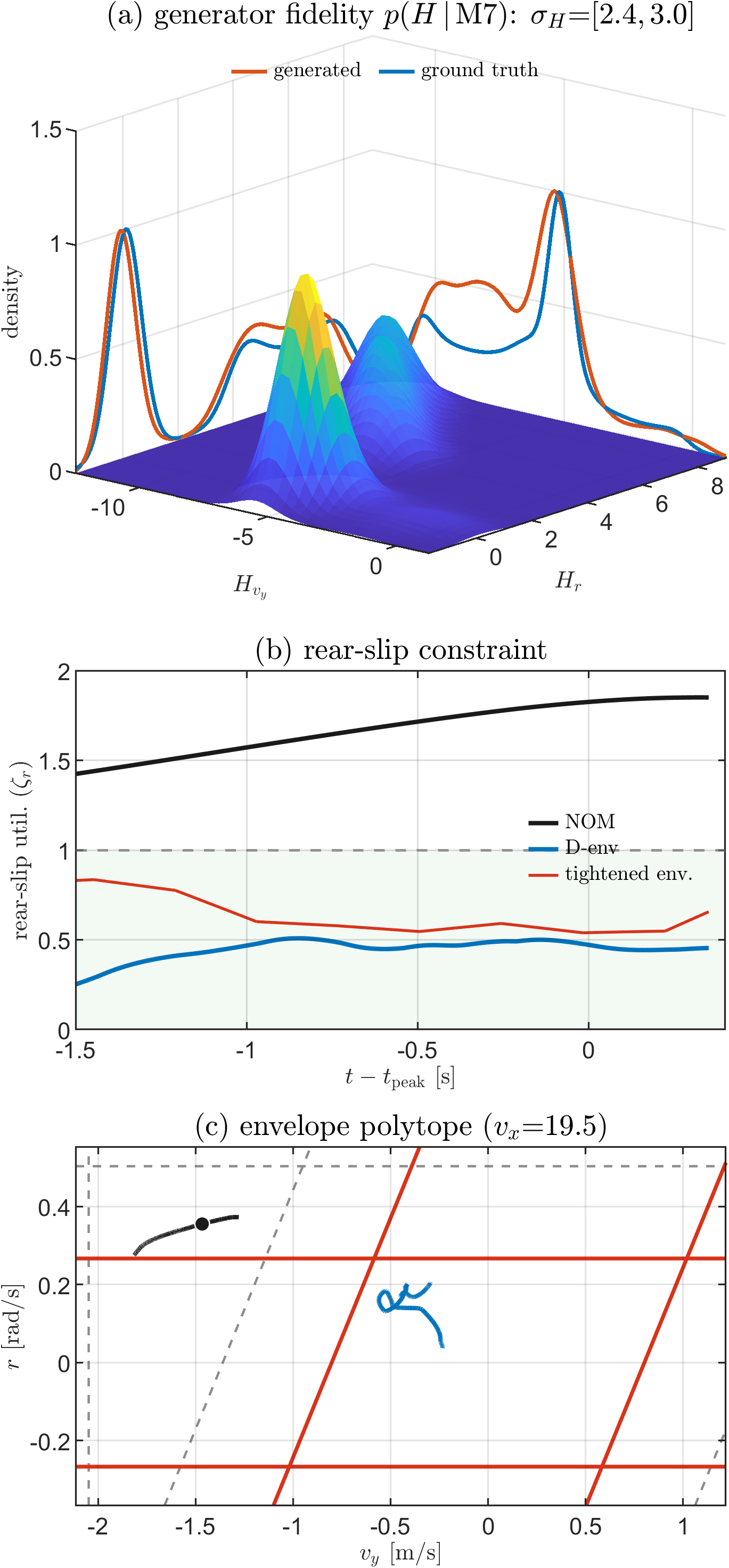}}{%
    \fbox{\parbox[c][0.95\columnwidth][c]{0.98\columnwidth}{\centering\itshape
  [fig\_dres\_bc.png pending --- run \texttt{gen\_m7\_7dof.m}, \texttt{gen\_m7\_traj.m}, then \texttt{build\_dres\_fig4.m}]}}}
  \caption{D-env on the M7-high brake-in-turn (sports car, 7-DOF). \textbf{(a)}~generated joint residual (oracle conditioning) vs.\ 7-DOF ground truth.
    \textbf{(b)}~rear-slip utilization $\zeta_r$ vs.\ time from the NOM side-slip peak $t_{\mathrm{peak}}$.
  \textbf{(c)}~the $(\vy,r)$ stable-handling envelope at the rear-slip-critical speed.}
  \label{fig:dres-3d}
\end{figure}

The fidelity of the generated $\sigma_H$ against a 7-DOF ground
truth~\cite{rajamani2012vehicle,pacejka2012tire} is verified in Fig.~\ref{fig:dres-3d}(a).
To isolate the generator's \emph{representational} accuracy, those draws are conditioned on the full
operating point---the state $x{=}[\vy\ r]^\top$ with the command, an oracle available only offline: at $90$
points along the M7-high brake-in-turn, $150$ draws track the ground-truth residual~\eqref{eq:residual} in
both center and spread. The generated \emph{spread}
$\sigma_H{=}[\sigma_{H_{\vy}},\sigma_{H_r}]{\approx}[2.4~\mathrm{m/s^2},\,3.0~\mathrm{rad/s^2}]$---a standard
deviation, hence positive---matches the ground truth $[2.4,\,3.1]$ within $3.1\%$; the residual mean is
separately offset, its $H_{\vy}$ center negative on this braking maneuver as shown in~Fig.~\ref{fig:dres-3d}(a).
This $3.1\%$ is an \emph{oracle} upper bound on representational capacity, however: conditioned on the command
alone ($[\vx,\delta]$, the input available online), the deployed model is markedly weaker---$\sigma_H\!\approx\![3.58,\,2.35]$
against the ground truth $[2.45,\,3.12]$, over-estimating $\sigma_{H_{\vy}}$ by $46\%$ and \emph{under}-estimating the
yaw spread $\sigma_{H_r}$ by $25\%$. The closed-loop tables of Section~\ref{sec:dres} use this command-only
generator throughout, so the reported gains already reflect deployed fidelity. The asymmetry is also
structurally contained: $\sigma_{H_r}$ enters the yaw bound only through its down-weighted state term
($k_1{=}0.25\,k_2$, $s_{3,k}{\approx}0.4$--$0.7\,\sigma_r$; Section~\ref{sub:dref}), so the $-25\%$ shifts
the total yaw back-off by only ${\approx}2$--$4\%$, while the $+46\%$ over-estimate pads the rear-slip
facet that binds in the brake-in-turn in the \mbox{conservative direction}.

\subsection{Chance-constrained stable-handling envelope via the projected residual}\label{sub:mpc}
D-env recasts the stable-handling
envelope~\eqref{eq:nmpc-beta}--\eqref{eq:nmpc-yaw} of the nominal controller~\eqref{eq:nominal-mpc} as a
single \emph{joint} chance constraint driven by the projected residual. To this end, the state residual
$\Hres$ of~\eqref{eq:residual} enters the predicted single-track dynamics~\eqref{eq:bicycle-d} as an
additive disturbance,
\begin{equation}\label{eq:dres-pred}
  x_{k+1}=A_d(\vx)\,x_k+B_d\,\delta_k+E_d\,\Hres_k,\quad E_d=T_s I_2 .
\end{equation}
On the controller's state plane $x=[\vy\ r]^\top$, the stable-handling envelope is the state polytope
\begin{equation}\label{eq:env-polytope}
  \mathcal{X}_{\mathrm{env}}(\vx)=\big\{\,x:\ |a_j^\top x|\le b_j(\vx),\ j=1,2,3\,\big\},
\end{equation}
where the three facets are the side-slip, rear-slip, and yaw-rate envelope
constraints~\eqref{eq:nmpc-beta}--\eqref{eq:nmpc-yaw}, respectively: $a_1\!\propto\![1,0]$, $a_2\!\propto\![1,-\ell_r]$, $a_3\!=\![0,1]$, and expand explicitly to
\begin{subequations}\label{eq:env-facets}
  \begin{align}
    |\vy|          &\le \vx\tan\beta_{\max},  &&\text{for } j{=}1,\label{eq:facet-beta}\\
    |\vy-\ell_r r| &\le \vx\alpha_{r,\max},   &&\text{for } j{=}2,\label{eq:facet-rslip}\\
    |r|            &\le \mu g/\vx,            &&\text{for } j{=}3.\label{eq:facet-yaw}
  \end{align}
\end{subequations}
Each facet is a pair of signed half-planes $\pm a_j^\top x\le b_j$ in the QP (six rows). Since the generator makes the predicted state a
distribution rather than a single point, D-env enforces this polytope probabilistically, requiring the
predicted state to lie in $\mathcal{X}_{\mathrm{env}}(\vx)$ with high
probability~\cite{carvalho2015stochastic,hewing2020cautious},
\begin{equation}\label{eq:joint-chance}
  \Pr\!\big[\,x_k\in\mathcal{X}_{\mathrm{env}}(\vx)\ \big|\ c\,\big]\;\ge\;1-\eta .
\end{equation}

Constraint~\eqref{eq:joint-chance} could be enforced directly on the $S$ propagated scenarios
$x_k^{(s)}$---a distribution-free scenario constraint with finite-sample
guarantees~\cite{calafiore2006scenario}---but for a single smooth QP D-env instead approximates it on each
facet separately. Under a Gaussian predicted state $x_k\mid c\sim\mathcal N(\bar x_k,\Sigma_{x,k})$ the projection
$a_j^\top x_k$ onto a facet normal is itself Gaussian, so each one-sided chance constraint
$\Pr[a_j^\top x_k\le b_j(\vx)\mid c]\ge 1-\eta$ admits the closed-form
tightening~\cite{calafiore2006distributionally,cannon2009probabilistic,mesbah2016stochastic} $a_j^\top\bar x_k\le
b_j(\vx)-\kappa\,s_{j,k}$, with $\kappa=\Phi^{-1}(1-\eta)$ and the facet-projected deviation
$s_{j,k}=\sqrt{a_j^\top\Sigma_{x,k}a_j}$. By Boole's inequality, enforcing each of the $m$ active facets at
this per-facet level $1-\eta$ delivers the joint envelope~\eqref{eq:joint-chance} at level at least
$1-m\eta$; in deployment a single facet is active (below), so the joint and per-facet levels coincide.
Adding a shared slack $\xi$ that keeps the QP feasible gives
the one-sided chance back-off
\begin{equation}\label{eq:joint-backoff}
  |a_j^\top\bar{x}_k|\;\le\;b_j(\vx)-\kappa\,s_{j,k}+\xi,
\end{equation}
where the facet deviation $s_{j,k}$ and the covariance $\Sigma_{x,k}$---the residual spread propagated
through~\eqref{eq:dres-pred} over the MPC's committed horizon---are
\begin{subequations}\label{eq:proj-cov}
  \begin{align}
    s_{j,k}&=\sqrt{a_j^\top\Sigma_{x,k}\,a_j},\quad \kappa=\Phi^{-1}(1-\eta),\label{eq:proj-cov-s}\\
    \Sigma_{x,k}&=G_k\,\mathrm{diag}(\sigma_H^2)\,G_k^\top,\quad
    G_k=\textstyle\sum_{i<k}A_d^{\,i}(\vx)\,T_s .\label{eq:proj-cov-Sigma}
  \end{align}
\end{subequations}
Equation~\eqref{eq:proj-cov-Sigma} is the linear covariance propagation of the disturbed
prediction~\eqref{eq:dres-pred}~\cite{mesbah2016stochastic,hewing2020cautious}: from a deterministic current
state the deviation accumulates the residual through the transition matrices,
\begin{equation}\label{eq:cov-accum}
  x_k-\bar x_k=\sum_{i<k}A_d^{\,i}(\vx)\,E_d\,\Hres=G_k\,\Hres,
\end{equation}
whose covariance for $\Hres\sim(0,\operatorname{diag}(\sigma_H^2))$ yields~\eqref{eq:proj-cov-Sigma}. Only
the \emph{centered} residual enters here: its nonzero mean is deliberately not propagated through the state
prediction but acts solely on the reference (Section~\ref{sub:dref}). The residual is
modeled as \emph{fully correlated} across the committed horizon rather than redrawn i.i.d.---a deliberately
conservative assumption consistent with a model-structure error that persists at a given operating point.
For the two axis-aligned facets the projection reduces to a single covariance entry,
\begin{equation}\label{eq:beta-yaw-proj}
  s_{1,k}=\sqrt{a_1^\top\Sigma_{x,k}\,a_1}=\sqrt{\Sigma_{\vy,k}}, \quad s_{3,k}=\sqrt{\Sigma_{r,k}}.
\end{equation}
Denoting the entries of
$\Sigma_{x,k}$ by $\Sigma_{\vy,k},\Sigma_{r,k}$ (diagonal) and $\Sigma_{\vy r,k}$ (off-diagonal), the
rear-slip projection $a_2=[1,-\ell_r]$ expands to
\begin{equation}\label{eq:rslip-proj}
  s_{2,k}=\sqrt{a_2^\top\Sigma_{x,k}\,a_2}
  =\sqrt{\Sigma_{\vy,k}-2\ell_r\,\Sigma_{\vy r,k}+\ell_r^2\,\Sigma_{r,k}},
\end{equation}
whose cross term $-2\ell_r\,\Sigma_{\vy r,k}$---the off-diagonal $\Sigma_{\vy r,k}$ generated by the
horizon propagation---is the $\vy$--$r$ coupling a yaw-only back-off discards, load-bearing only in the
rear-slip-critical brake-in-turn regime (M7, Fig.~\ref{fig:dres-3d}) and contributing $<\!0.05^\circ$
elsewhere across the evaluated sweep.
The deployed controller applies a common one-sided level $\kappa=\Phi^{-1}(0.95)=1.645$ to every active
facet, softened by a shared slack $\xi$ (penalized by $w_\xi\xi^2$
in~\eqref{eq:nmpc-cost}) so the QP remains feasible: an over-tight back-off relaxes into a small, penalized
violation rather than an infeasible program, so the controller degrades gracefully instead of failing.
Across the evaluated maneuvers only the friction-limited yaw facet~\eqref{eq:facet-yaw} is consistently
active, so in deployment the multi-facet joint back-off reduces to the scalar yaw term
$s_{3,k}=\sqrt{\Sigma_{r,k}}$; the full form of~\eqref{eq:joint-backoff} is the general mechanism, whose
rear-slip cross term is exercised only in the brake-in-turn regime (M7, Fig.~\ref{fig:dres-3d}). This
soft relaxation gives \emph{persistent} QP feasibility by construction, not a set-invariance guarantee;
recursive feasibility and closed-loop stability, established here empirically across the maneuver and
vehicle sweep, are left to future work rather than to a terminal-cost or invariant-set certificate.

Figure~\ref{fig:dres-3d} applies D-env to the high-$\mu$ brake-in-turn (M7-high), the maneuver where the
rear-slip back-off is most active---marginal for the nominal controller on the design sedan
(Fig.~\ref{fig:traj-featured}) but a \emph{violation} on the sports car shown here:
the rear-slip utilization $\zeta_r$ is plotted against its chance-tightened bound in Fig.~\ref{fig:dres-3d}(b),
and the same trajectory in the $(\vy,r)$ state plane at the rear-slip-critical speed in Fig.~\ref{fig:dres-3d}(c).
The rear-slip utilization is defined as
\begin{equation}\label{eq:rslip-util}
  \zeta_r \;=\; \frac{|\vy-\ell_r r|}{\vx\,\alpha_{r,\max}},
\end{equation}
and is shown against the chance-tightened rear-slip bound---the rear-slip facet of the joint
back-off~\eqref{eq:joint-backoff} in the same utilization units, floored at
$\alpha_{r,\min}/\alpha_{r,\max}$ ($\alpha_{r,\min}{=}1^\circ$) so the corridor stays open under large
spread,
\begin{equation}\label{eq:rslip-tight}
  \zeta_r^{\mathrm{tight}} \;=\; \max\!\Big(\tfrac{\alpha_{r,\min}}{\alpha_{r,\max}},\;
  1-\frac{\kappa\,s_{2,k}}{\vx\,\alpha_{r,\max}}\Big).
\end{equation}
In Fig.~\ref{fig:dres-3d}(b), time is referenced to $t_{\mathrm{peak}}$, the NOM side-slip peak, and
$\zeta_r\!>\!1$ is a rear-slip violation~\eqref{eq:nmpc-rslip}. NOM overruns the bound by
$1.85\times$ as its side-slip diverges to $\beta_{\mathrm{peak}}\!=\!6.07^\circ$, past the $6^\circ$ envelope---the
$^{\times}$ entry for this sports-car M7-high cell of Table~\ref{tab:cmp-7dof}. D-env holds $\zeta_r$ below
the tightened bound throughout and completes at $\beta_{\mathrm{peak}}\!=\!1.65^\circ$. In Fig.~\ref{fig:dres-3d}(c) the rear-slip
line~\eqref{eq:nmpc-rslip} is one edge of the Beal--Gerdes
envelope: NOM crosses it, while D-env stays within the chance-tightened polytope whose
rear-slip edge is pulled in by $\sqrt{a_2^\top\Sigma_{x,k}\,a_2}$---the off-diagonal of $\Sigma_{x,k}$, the
propagation-induced $\vy$--$r$ coupling that a yaw-only back-off ignores.

\subsection{Diffusion-reference MPC (D-ref)}\label{sub:dref}
D-ref is the \emph{second} use of the same generator (Section~\ref{sub:residual}). Where D-env
(Section~\ref{sub:mpc}) tightened the stable-handling envelope with the propagated \emph{state}-residual
spread, D-ref reads the \emph{reference} residual on the same horizon and uses \emph{both} its moments at
once: the \emph{mean} re-sizes the tracked yaw reference, and the \emph{spread} additionally backs off the
friction-limited yaw bound. The reference residual is the scalar gap between the achievable yaw rate
$r$ and the linear yaw reference $r^{\mathrm{lin}}\!$ in~\eqref{eq:rref},
\begin{equation}\label{eq:refresidual}
  \Delta r^{\mathrm{ref}} \;=\; r-r^{\mathrm{lin}},
\end{equation}
conditioned on the command $c=[\vx,\delta]$ and learned to absorb the understeer-gradient error, tire
nonlinearity, and parameter/road mismatch near the limit. Its two moments,
\begin{equation}\label{eq:refmoments}
  \Delta\bar{r}^{\mathrm{ref}}=\mathbb{E}[\Delta r^{\mathrm{ref}}\mid c],\qquad
  \sigma_r=\operatorname{std}[\Delta r^{\mathrm{ref}}\mid c],
\end{equation}
are estimated from $S$ independent reverse-process draws: the mean $\Delta\bar{r}^{\mathrm{ref}}$ re-sizes the reference and the
spread $\sigma_r$ drives the yaw-bound back-off. The count $S{=}64$
is itself a design choice. We verify it by rebuilding the deployed $\sigma_r$ surface at varying $S$ and
re-running the 7-DOF closed loop (three-seed mean): in this convergence study the pooled fleet-median
$\beta_{\mathrm{peak}}$ falls from $1.16^\circ$ at
$S{=}16$ (noisy, $\pm0.35^\circ$ three-seed spread on M7) to $1.03^\circ$ at $S{=}64$ and essentially no
lower above it ($1.02^\circ$~at~$S{\ge}128$).

Near the limit the tires saturate, so the steady-state understeer relation~\eqref{eq:rref} over-predicts
the achievable yaw (Section~\ref{sub:afs})---increasingly on off-nominal vehicles (Fig.~\ref{fig:diversity}).
Tracking this inflated reference would breach the very stability limit the controller enforces. D-ref
corrects it by adding the conditional mean of the reference residual $\Delta r^{\mathrm{ref}}$ to the linear
prediction, giving the \emph{achievable} reference the nominal LTV-MPC~\eqref{eq:nominal-mpc} then tracks,
\begin{subequations}\label{eq:diffref}
  \begin{align}
    r^{\mathrm{ref}}_k&=r^{\mathrm{lin}}_k+\mathbb{E}\big[\Delta r^{\mathrm{ref}}\mid c_k\big], \label{eq:diffref-gen}\\
    |r^{\mathrm{ref}}_k|&\le \mu g/\vx . \label{eq:diffref-clamp}
  \end{align}
\end{subequations}
The mean correction~\eqref{eq:diffref-gen} re-sizes the reference and the bound~\eqref{eq:diffref-clamp}
keeps it inside the friction limit. Because $\Delta r^{\mathrm{ref}}$ is the gap to the linear
reference~\eqref{eq:refresidual}, its conditional mean is exactly that reference's bias, so
$r^{\mathrm{lin}}_k+\Delta\bar{r}^{\mathrm{ref}}=\mathbb{E}[r\mid c_k]$ is the unbiased (minimum-mean-square-error) estimate of the
achievable yaw. The mean thus enters with \emph{unit weight}: all conservatism is provided by the
\emph{separate} spread back-off $\sigma_r$ below, and a sub-unit gain on $\Delta\bar{r}^{\mathrm{ref}}$ would only re-introduce the
understeer bias D-ref removes. Only the reference is corrected by a mean---feeding the \emph{state}
residual's mean forward would overlap the robust feedback that already rejects it (a worse estimate then
absorbed by higher gain, amplifying noise and chattering the actuator~\cite{dronediffusion2025}), whereas
re-sizing the tracked reference acts on the predictive layer alone. The achievable yaw is, however, a
learned \emph{distribution}, not a point. Realizations of $\Delta r^{\mathrm{ref}}$ scatter by $\sigma_r$~\eqref{eq:refmoments} about $\Delta\bar{r}^{\mathrm{ref}}$,
so a mean-feasible reference can still demand more yaw than the vehicle can develop---saturating the
tracker and breaching the stability limit that the bound encodes. We therefore enforce the friction-limited yaw bound as a \emph{chance constraint},
\begin{equation}\label{eq:yaw-chance}
  \Pr\!\big[\,|r_k|\le \mu g/\vx \,\big|\, c_k\,\big]\;\ge\;1-\eta,
\end{equation}
exposed to the two uncertainties the generator quantifies: the achievable-yaw reference scatters by
$\sigma_r$~\eqref{eq:refmoments} (D-ref), while the propagated state residual gives the predicted yaw a
spread $s_{3,k}=\sqrt{\Sigma_{r,k}}$~\eqref{eq:beta-yaw-proj} (D-env, Section~\ref{sub:mpc}). Treating both
as zero-mean Gaussian perturbations of $r_k$ about its mean, and noting that only the active half-plane of
the yaw bound binds at the limit, we apportion the risk $\eta=\eta_1+\eta_2$ across the two sources; Boole's
inequality then bounds the one-sided overshoot (WLOG the upper half-plane) by the sum of their one-sided
tails~\cite{mesbah2016stochastic,carvalho2015stochastic,hewing2020cautious},
\begin{equation}\label{eq:boole}
  \begin{aligned}
    \Pr\!\big[\,r_k>\mu g/\vx\,\big] &\;\le\; \Phi(-k_1)+\Phi(-k_2)\\
    &\;=\; \eta_1+\eta_2 \;=\; \eta,
  \end{aligned}
\end{equation}
where $k_i=\Phi^{-1}(1-\eta_i)$.
Subtracting both margins from the friction limit yields the joint yaw bound,
\begin{equation}\label{eq:yaw-joint}
  |r_k|\;\le\;\frac{\mu g}{\vx}-\big(k_1 s_{3,k}+k_2\,\sigma_r\big)+\xi .
\end{equation}
The split~\eqref{eq:boole} is a risk \emph{allocation}: a violation implies some perturbation exceeds its
own margin, so the bound holds under \emph{arbitrary} dependence---appropriate here, since both terms stem
from the same grip deficit and are not independent---at the price of conservatism over the combined-variance
back-off
$\kappa\sqrt{s_{3,k}^2+\sigma_r^2}$ independence would permit. The allocation $k_1{=}0.25\,k_2$ down-weights
the small state term relative to the reference term rather than equalizing the split~\eqref{eq:boole}; it is
tuned to the observed $s_{3,k}\!\approx\!0.4$--$0.7\,\sigma_r$, not derived. The reference term carries the
nominal one-sided level ($k_2{=}\Phi^{-1}(0.95){=}1.645$, $\eta_2{=}0.05$) and dominates the bound, the
state term adding a smaller, deliberately conservative margin.
Since it peaks near the friction
limit the bound tightens precisely where the achievable yaw is least certain and becomes negligible in the
linear regime. D-ref thus acts as a \emph{learned} reference governor~\cite{garone2017reference}: it never commands a yaw
reference beyond the feasible set, with a margin set by the data-driven $\sigma_r$ rather than a fixed rule.

The reference enters only as an \emph{exogenous} target, not propagated through the prediction dynamics like
the D-env state residual (Section~\ref{sub:mpc}). We condition it on the command alone: since
$r^{\mathrm{ref}}\!\approx\!r$, conditioning on $r$ (or $a_y\!\approx\!\vx r$) would let the generator recover
its own target directly from the input. The road friction $\mu$ is likewise \emph{not} a conditioning input,
and the residuals are pooled across the dry and low-$\mu$ training conditions, so $\sigma_r$~\eqref{eq:refmoments}
is a command-conditioned spread \emph{marginalized} over the training surface mix rather than a per-surface
predictive uncertainty; the $\mu$-dependent yaw ceiling is instead enforced by the hard
clamp~\eqref{eq:diffref-clamp} and the friction-limited bound~\eqref{eq:yaw-joint}. The pooling is a
deliberate trade. On dry roads the marginal spread over-covers, costing only the modest path margin the
back-off trades anyway (below); on low-$\mu$ roads, where it under-states the per-surface spread, the
recovery is carried by the mean re-size, aided by the measured-utilization
rescaling~\eqref{eq:nmpc-stiff}. Conditioning on an online $\hat\mu$ would instead tie the margins to a
friction classifier whose misclassification at the limit would step-change the constraint set; the pooled
spread degrades gracefully.

\begin{figure}[t]
  \centering
  \IfFileExists{fig_dref_map.png}{%
  \includegraphics[width=0.80\columnwidth]{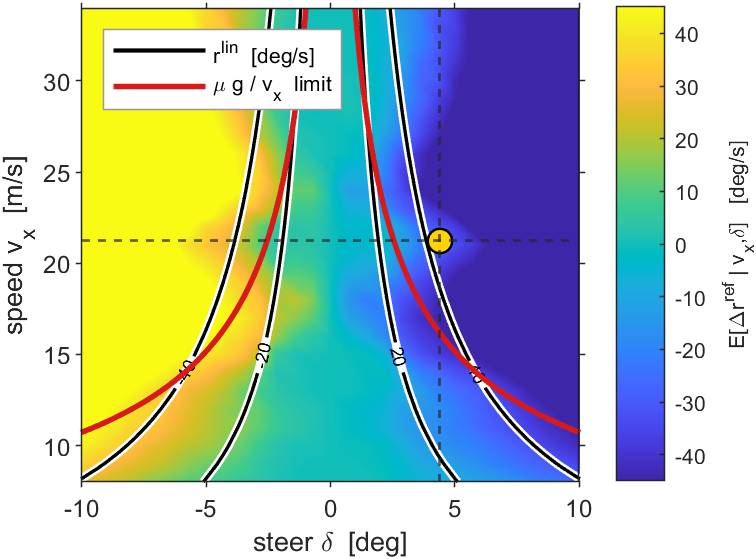}}{%
    \fbox{\parbox[c][0.55\columnwidth][c]{0.95\columnwidth}{\centering\itshape
        [fig\_dref\_map.png pending --- run \texttt{gen\_dref\_map.m} then
  \texttt{fig\_dref\_map\_m.m}]}}}
  \caption{Diffusion reference correction $\mathbb{E}[\Delta r^{\mathrm{ref}}\mid\vx,\delta]$ over the
    command plane (7-DOF sedan); the marked point is the M3-step operating point, where the linear
  reference over-predicts the achievable yaw and the diffusion mean re-sizes it.}
  \label{fig:dref-map}
\end{figure}

Unlike a diffusion planner that generates the executed
trajectory~\cite{janner2022diffuser,ajay2023decisiondiffuser}, D-ref generates only the tracked reference.
The learned correction $\mathbb{E}[\Delta r^{\mathrm{ref}}\mid\vx,\delta]$ over the command plane is shown in Fig.~\ref{fig:dref-map}.
The marked point in the figure is an M3-step operating point at $\vx{=}21.2$\,m/s, $\delta{=}4.4^\circ$, where the linear reference over-predicts
at $r^{\mathrm{lin}}{=}45.7^\circ$/s. There the mean correction
$\mathbb{E}[\Delta r^{\mathrm{ref}}]{=}-29.2^\circ$/s re-sizes it to $16.5^\circ$/s---the achievable yaw,
itself well inside the friction limit $\mu g/\vx{=}26.5^\circ$/s.
The deployed controller carries a fixed nominal $\mu{=}1$ and does not observe the road surface, so the
friction clamp~\eqref{eq:diffref-clamp}, the envelope yaw facet~\eqref{eq:facet-yaw}, and the friction-circle
stiffness rescaling~\eqref{eq:nmpc-stiff} are all $\mu{=}1$-anchored. On a low-$\mu$ surface these bounds are
therefore loose and typically inactive; the recovery there is carried by the learned \emph{mean} re-size
$\mathbb{E}[\Delta r^{\mathrm{ref}}]$, which absorbs the grip deficit through the command-conditioned
residual, as the closed-loop low-$\mu$ results confirm (Section~\ref{sec:dres}).

In closed loop this re-sizing lowers $\beta_{\mathrm{peak}}$ without sacrificing path-following. D-ref holds the
lateral path deviation at essentially the NOM level (fleet-median devMax $2.0$ vs.\ $2.1$\,m), while the
deeper reductions of the full D-res arise from the chance back-off, which trades a modest path margin
($2.1\!\to\!2.6$\,m, the path-dev row of Table~\ref{tab:cmp-7dof}) for the additional~stability.

These command-conditioned moments are tabulated offline, so D-res deploys with \emph{no in-loop
diffusion}: unlike diffusion planners that denoise inside the control loop, no generative inference runs at
the control rate. The only per-step cost over the nominal MPC is one $O(1)$ bilinear lookup of the
moments~\eqref{eq:refmoments},~\eqref{eq:sigmaH}---made once at the current command and held over the
committed horizon---plus the analytic envelope back-off~\eqref{eq:joint-backoff}
formed from $\Sigma_{x,k}$~\eqref{eq:proj-cov}, both negligible against the fixed-iteration QP that
dominates the step. The interpolated moment surfaces are smooth (Fig.~\ref{fig:dref-map}) and enter only
the reference and the constraint right-hand sides, which bilinear interpolation keeps continuous across
grid cells; no steering chattering was observed over the evaluated sweep. On the embedded target the full
D-res controller meets the $100$\,Hz budget with margin to spare (Section~\ref{sub:realtime}), so D-res
both inherits and demonstrates the baseline LTV-MPC's real-time~feasibility.

%% file: sections/05-experiments.tex
\section{Comparison and Ablation}\label{sec:dres}

Directly comparing D-res against the many limit-handling controllers surveyed in
Section~\ref{sec:related} (Table~\ref{tab:priorwork}) would be the ideal benchmark, but their
implementations are not uniformly public, which precludes an exact, reproducible comparison. We therefore
anchor the study on the deliberately simple nominal controller~\eqref{eq:nominal-mpc} and perform a
controlled ablation, adding the envelope back-off (D-env) and then the learned reference (GP-ref and
D-res) one ingredient at a time. The evaluation is split into two tiers---a 7-DOF vehicle model
(Section~\ref{sub:eval-7dof}) and high-fidelity CarMaker~13 (Section~\ref{sub:cmp-cm})---which, following
the driver and vehicle axes of Sections~\ref{sub:afs} and~\ref{sub:vehicle}, test whether the \emph{same}
controller holds under different driver models and vehicle dynamics; the diffusion generator is retrained
per tier. Finally, Section~\ref{sub:abl-diversity} probes the robustness of the controller to load and
tire variation.

\subsection{Evaluation setup}\label{sub:setup}

\begin{table*}[p]
  \centering
  \rotatebox{90}{%
  \begin{minipage}{0.92\textheight}
  \centering
  \captionof{table}{Ablation on the 7-DOF plant: closed-loop $\beta_{\mathrm{peak}}$ [deg]; KPIs and markers
  ($^{\times}$, path dev.). Reduction percentages are computed from the printed medians.}
  \label{tab:cmp-7dof}
  \renewcommand{\arraystretch}{1.08}
  \setlength{\tabcolsep}{3pt}
  \footnotesize
  \resizebox{\linewidth}{!}{%
    \begin{tabular}{@{}ll *{20}{c}@{}}
      \toprule
      & & \multicolumn{5}{c}{Sedan} & \multicolumn{5}{c}{Compact} & \multicolumn{5}{c}{Sports} & \multicolumn{5}{c}{SUV}\\
      \cmidrule(lr){3-7}\cmidrule(lr){8-12}\cmidrule(lr){13-17}\cmidrule(lr){18-22}
      Maneuver & $\mu$ & NOM & D-env & D-ref & GP-ref & D-res & NOM & D-env & D-ref & GP-ref & D-res & NOM & D-env & D-ref & GP-ref & D-res & NOM & D-env & D-ref & GP-ref & D-res\\
      \midrule
      M1 DLC           & high & $3.27$ & $2.85$ & $3.06$ & $2.99$ & $\mathbf{2.81}$ & $3.49$ & $\mathbf{2.72}$ & $3.48$ & $2.96$ & $\mathbf{2.72}$ & $3.53$ & $3.21$ & $3.60$ & $3.10$ & $\mathbf{2.88}$ & $3.27$ & $\mathbf{2.44}$ & $3.37$ & $2.60$ & $2.47$\\
                       & low  & $2.66$ & $4.85$ & $3.07$ & $\mathbf{1.50}$ & $2.42$ & $2.73$ & $2.73$ & $\mathbf{2.53}$ & $2.72$ & $2.69$ & $3.20$ & $3.21$ & $3.20$ & $\mathbf{1.73}$ & $2.35$ & $\mathbf{2.43}$ & $2.53$ & $2.95$ & $3.34$ & $2.59$\\
      M2 severe DLC    & high & $2.48$ & $1.43$ & $\mathbf{0.95}$ & $1.42$ & $0.96$ & $3.37$ & $1.89$ & $1.85$ & $1.82$ & $\mathbf{1.65}$ & $2.84$ & $2.04$ & $\mathbf{1.08}$ & $1.51$ & $\mathbf{1.08}$ & $2.70$ & $1.57$ & $1.31$ & $1.44$ & $\mathbf{1.21}$\\
      M3 step          & high & $2.11$ & $0.76$ & $1.49$ & $0.98$ & $\mathbf{0.75}$ & $0.98$ & $\mathbf{0.41}$ & $0.82$ & $1.58$ & $\mathbf{0.41}$ & $3.16$ & $0.25$ & $0.55$ & $\mathbf{0.13}$ & $0.22$ & $1.03$ & $0.72$ & $1.20$ & $\mathbf{0.66}$ & $0.72$\\
                       & low  & $2.41$ & $1.97$ & $1.25$ & $\mathbf{0.33}$ & $1.25$ & $2.02$ & $2.13$ & $\mathbf{0.96}$ & $1.95$ & $\mathbf{0.96}$ & $3.54$ & $2.57$ & $1.17$ & $\mathbf{0.31}$ & $1.17$ & $1.76$ & $1.48$ & $\mathbf{1.35}$ & $1.39$ & $1.46$\\
      M4 circular      & high & $1.16$ & $1.16$ & $\mathbf{1.08}$ & $1.09$ & $\mathbf{1.08}$ & $1.56$ & $1.56$ & $\mathbf{1.32}$ & $1.52$ & $\mathbf{1.32}$ & $1.19$ & $1.19$ & $\mathbf{1.09}$ & $1.13$ & $\mathbf{1.09}$ & $1.24$ & $1.24$ & $1.05$ & $\mathbf{0.94}$ & $1.05$\\
      M5 sine-with-dwell & high & $3.60$ & $0.91$ & $2.11$ & $\mathbf{0.85}$ & $0.87$ & $3.72$ & $0.96$ & $2.29$ & $1.59$ & $\mathbf{0.93}$ & $3.76$ & $0.66$ & $1.38$ & $1.06$ & $\mathbf{0.63}$ & $3.55$ & $\mathbf{0.94}$ & $2.21$ & $1.09$ & $\mathbf{0.94}$\\
      M6 sweep         & high & $0.81$ & $0.81$ & $\mathbf{0.52}$ & $1.37$ & $\mathbf{0.52}$ & $0.52$ & $0.52$ & $0.41$ & $\mathbf{0.38}$ & $0.41$ & $1.24$ & $1.01$ & $\mathbf{0.79}$ & $1.41$ & $\mathbf{0.79}$ & $0.48$ & $0.48$ & $0.59$ & $\mathbf{0.40}$ & $0.59$\\
                       & low  & $0.83$ & $0.83$ & $\mathbf{0.53}$ & $0.75$ & $\mathbf{0.53}$ & $0.95$ & $0.95$ & $\mathbf{0.38}$ & $0.91$ & $\mathbf{0.38}$ & $1.12$ & $1.12$ & $\mathbf{0.59}$ & $0.85$ & $\mathbf{0.59}$ & $0.76$ & $0.76$ & $\mathbf{0.55}$ & $0.87$ & $\mathbf{0.55}$\\
      M7 brake-in-turn & high & $3.66$ & $1.97$ & $2.37$ & $\mathbf{1.70}$ & $1.88$ & $2.16$ & $2.10$ & $2.23$ & $\mathbf{1.82}$ & $2.14$ & $6.07^{\times}$ & $1.65$ & $3.47$ & $2.52$ & $\mathbf{1.50}$ & $1.79$ & $1.56$ & $1.98$ & $\mathbf{1.32}$ & $1.77$\\
      \midrule
      \textit{Fleet median} ($\beta_{\mathrm{peak}}$) [deg] &
      & \shortstack{$2.45$\\$-$} & \shortstack{$1.29$\\$\downarrow47\%$} & \shortstack{$1.37$\\$\downarrow44\%$} & \shortstack{$1.23$\\$\downarrow50\%$} & \shortstack{$\mathbf{1.02}$\\$\downarrow58\%$}
      & \shortstack{$2.09$\\$-$} & \shortstack{$1.72$\\$\downarrow18\%$} & \shortstack{$1.59$\\$\downarrow24\%$} & \shortstack{$1.70$\\$\downarrow19\%$} & \shortstack{$\mathbf{1.14}$\\$\downarrow45\%$}
      & \shortstack{$3.18$\\$-$} & \shortstack{$1.42$\\$\downarrow55\%$} & \shortstack{$1.13$\\$\downarrow64\%$} & \shortstack{$1.27$\\$\downarrow60\%$} & \shortstack{$\mathbf{1.09}$\\$\downarrow66\%$}
      & \shortstack{$1.78$\\$-$} & \shortstack{$1.36$\\$\downarrow23\%$} & \shortstack{$1.33$\\$\downarrow25\%$} & \shortstack{$1.21$\\$\downarrow32\%$} & \shortstack{$\mathbf{1.13}$\\$\downarrow36\%$}\\
      \textit{Fleet max} ($\beta_{\mathrm{peak}}$) [deg] & & $3.66$ & $4.85$ & $3.07$ & $2.99$ & $\mathbf{2.81}$ & $3.72$ & $2.73$ & $3.48$ & $2.96$ & $\mathbf{2.72}$ & $6.07$ & $3.21$ & $3.60$ & $3.10$ & $\mathbf{2.88}$ & $3.55$ & $\mathbf{2.53}$ & $3.37$ & $3.34$ & $2.59$\\
      \midrule
      path dev.\ (M1,M2) [m] & & $\mathbf{2.01}$ & $2.39$ & $2.05$ & $3.16$ & $2.60$ & $2.07$ & $2.59$ & $\mathbf{2.02}$ & $2.43$ & $2.70$ & $2.13$ & $3.00$ & $\mathbf{2.08}$ & $3.21$ & $3.08$ & $2.08$ & $2.37$ & $\mathbf{1.98}$ & $2.85$ & $2.36$\\
      \bottomrule
  \end{tabular}}

  \vspace{1.5em}

  \captionof{table}{Ablation on the \emph{CarMaker}~13 tier: peak side-slip $\beta_{\mathrm{peak}}$ [deg]; KPIs, $^{\times}$,
    and spin per Section~\ref{sub:setup} ($^{\dagger}$: M6-low maneuver-design limit, excluded from the
  envelope-pass base, the fleet median, the fleet max, and the spin count).}
  \label{tab:cmp-cm}
  \renewcommand{\arraystretch}{1.08}
  \setlength{\tabcolsep}{3pt}
  \footnotesize
  \resizebox{\linewidth}{!}{%
    \begin{tabular}{@{}ll *{20}{c}@{}}
      \toprule
      & & \multicolumn{5}{c}{Sedan} & \multicolumn{5}{c}{Compact} & \multicolumn{5}{c}{Sports} & \multicolumn{5}{c}{SUV}\\
      \cmidrule(lr){3-7}\cmidrule(lr){8-12}\cmidrule(lr){13-17}\cmidrule(lr){18-22}
      Maneuver & $\mu$ & NOM & D-env & D-ref & GP-ref & D-res & NOM & D-env & D-ref & GP-ref & D-res & NOM & D-env & D-ref & GP-ref & D-res & NOM & D-env & D-ref & GP-ref & D-res\\
      \midrule
      M1 DLC           & high & $1.54$ & $1.15$ & $1.18$ & $\mathbf{0.62}$ & $0.72$ & $2.69$ & $1.06$ & $1.84$ & $1.57$ & $\mathbf{0.75}$ & $2.23$ & $2.18$ & $0.43$ & $0.62$ & $\mathbf{0.33}$ & $1.10$ & $\mathbf{0.40}$ & $0.77$ & $0.58$ & $0.53$\\
                       & low  & $7.87^{\times}$ & $14.54^{\times}$ & $1.60$ & $1.60$ & $\mathbf{0.25}$ & $4.63$ & $4.63$ & $\mathbf{0.15}$ & $3.73$ & $\mathbf{0.15}$ & $3.18$ & $2.86$ & $\mathbf{0.20}$ & $0.67$ & $\mathbf{0.20}$ & $3.16$ & $2.74$ & $0.27$ & $\mathbf{0.26}$ & $0.27$\\
      M2 severe DLC    & high & $0.82$ & $3.46$ & $1.55$ & $\mathbf{0.58}$ & $1.55$ & $1.11$ & $1.54$ & $0.56$ & $0.94$ & $\mathbf{0.45}$ & $2.18$ & $2.09$ & $\mathbf{0.35}$ & $2.94$ & $\mathbf{0.35}$ & $2.00$ & $6.13^{\times}$ & $\mathbf{1.78}$ & $1.93$ & $\mathbf{1.78}$\\
      M3 step          & high & $2.83$ & $2.21$ & $0.60$ & $0.60$ & $\mathbf{0.57}$ & $2.71$ & $0.77$ & $1.43$ & $1.64$ & $\mathbf{0.74}$ & $1.72$ & $1.17$ & $0.43$ & $2.85$ & $\mathbf{0.38}$ & $0.79$ & $0.51$ & $0.71$ & $\mathbf{0.46}$ & $0.49$\\
                       & low  & $99.54^{\times}$ & $109.94^{\times}$ & $0.46$ & $\mathbf{0.44}$ & $0.46$ & $12.87^{\times}$ & $12.87^{\times}$ & $\mathbf{0.19}$ & $6.80^{\times}$ & $\mathbf{0.19}$ & $124.79^{\times}$ & $109.81^{\times}$ & $\mathbf{0.29}$ & $7.49^{\times}$ & $\mathbf{0.29}$ & $109.50^{\times}$ & $109.15^{\times}$ & $\mathbf{0.63}$ & $113.16^{\times}$ & $\mathbf{0.63}$\\
      M4 circular      & high & $1.13$ & $1.11$ & $1.62$ & $1.62$ & $\mathbf{0.78}$ & $2.01$ & $1.93$ & $1.23$ & $1.58$ & $\mathbf{0.81}$ & $4.73$ & $4.64$ & $\mathbf{0.67}$ & $1.78$ & $\mathbf{0.67}$ & $5.09$ & $5.05$ & $\mathbf{0.90}$ & $4.88$ & $\mathbf{0.90}$\\
      M5 sine-with-dwell & high & $2.64$ & $1.19$ & $1.17$ & $0.60$ & $\mathbf{0.57}$ & $3.28$ & $0.74$ & $2.95$ & $1.78$ & $\mathbf{0.69}$ & $2.18$ & $1.48$ & $0.54$ & $3.16$ & $\mathbf{0.31}$ & $1.81$ & $\mathbf{0.53}$ & $1.10$ & $0.68$ & $0.54$\\
      M6 sweep         & high & $3.52$ & $0.78$ & $0.91$ & $0.91$ & $\mathbf{0.76}$ & $2.84$ & $\mathbf{1.21}$ & $2.13$ & $1.69$ & $1.33$ & $3.73$ & $\mathbf{0.60}$ & $1.47$ & $3.43$ & $\mathbf{0.60}$ & $3.32$ & $0.72$ & $0.99$ & $0.74$ & $\mathbf{0.70}$\\
                       & low  & $147.57^{\dagger}$ & $148.31^{\dagger}$ & $91.05^{\dagger}$ & $131.34^{\dagger}$ & $129.43^{\dagger}$ & $6.37^{\dagger}$ & $6.37^{\dagger}$ & $8.54^{\dagger}$ & $6.14^{\dagger}$ & $8.54^{\dagger}$ & $97.57^{\dagger}$ & $164.13^{\dagger}$ & $137.27^{\dagger}$ & $174.35^{\dagger}$ & $2.42^{\dagger}$ & $123.80^{\dagger}$ & $123.80^{\dagger}$ & $78.49^{\dagger}$ & $127.78^{\dagger}$ & $107.10^{\dagger}$\\
      M7 brake-in-turn & high & $2.95$ & $0.56$ & $0.81$ & $0.68$ & $\mathbf{0.43}$ & $2.65$ & $1.68$ & $2.47$ & $1.15$ & $\mathbf{0.56}$ & $1.78$ & $1.71$ & $\mathbf{0.38}$ & $1.79$ & $\mathbf{0.38}$ & $1.07$ & $\mathbf{0.54}$ & $1.32$ & $2.62$ & $0.56$\\
      \midrule
      \textit{Fleet median} ($\beta_{\mathrm{peak}}$) [deg] &
      & \shortstack{$2.83$\\$-$} & \shortstack{$1.19$\\$\downarrow58\%$} & \shortstack{$1.17$\\$\downarrow59\%$} & \shortstack{$0.62$\\$\downarrow78\%$} & \shortstack{$\mathbf{0.57}$\\$\downarrow80\%$}
      & \shortstack{$2.71$\\$-$} & \shortstack{$1.54$\\$\downarrow43\%$} & \shortstack{$1.43$\\$\downarrow47\%$} & \shortstack{$1.64$\\$\downarrow39\%$} & \shortstack{$\mathbf{0.69}$\\$\downarrow75\%$}
      & \shortstack{$2.23$\\$-$} & \shortstack{$2.09$\\$\downarrow6\%$} & \shortstack{$0.43$\\$\downarrow81\%$} & \shortstack{$2.85$\\$\uparrow28\%$} & \shortstack{$\mathbf{0.35}$\\$\downarrow84\%$}
      & \shortstack{$2.00$\\$-$} & \shortstack{$0.72$\\$\downarrow64\%$} & \shortstack{$0.90$\\$\downarrow55\%$} & \shortstack{$0.74$\\$\downarrow63\%$} & \shortstack{$\mathbf{0.56}$\\$\downarrow72\%$}\\
      \textit{Fleet max} ($\beta_{\mathrm{peak}}$) [deg] & & $99.54$ & $109.94$ & $1.62$ & $1.62$ & $\mathbf{1.55}$ & $12.87$ & $12.87$ & $2.95$ & $6.80$ & $\mathbf{1.33}$ & $124.79$ & $109.81$ & $1.47$ & $7.49$ & $\mathbf{0.67}$ & $109.50$ & $109.15$ & $\mathbf{1.78}$ & $113.16$ & $\mathbf{1.78}$\\
      \midrule
      \emph{spins} ($\beta_{\mathrm{peak}}\!\ge\!15^\circ$) & & $1$ & $1$ & $0$ & $0$ & $0$ & $0$ & $0$ & $0$ & $0$ & $0$ & $1$ & $1$ & $0$ & $0$ & $0$ & $1$ & $1$ & $0$ & $1$ & $0$\\
      \bottomrule
  \end{tabular}}
  \end{minipage}}
\end{table*}
On both evaluation tiers the four-vehicle fleet (Table~\ref{tab:fleet}) is controlled with one fixed
single-track parameter set over the standardized ISO/FMVSS handling tests of
Table~\ref{tab:scenarios}. Every maneuver is scored by one common \emph{stability} key performance
indicator (KPI):
the peak side-slip $\beta_{\mathrm{peak}}$ [deg] over the moving window ($\vx\!>\!5$\,m/s), the canonical
loss-of-control indicator~\cite{inagaki1994analysis,vanzanten2000esp}. The pass criterion is the
$\pm6^\circ$ stable-handling envelope---a conservative margin well inside the $\sim$10--12$^\circ$
front-steering yaw-authority limit~\cite{inagaki1994analysis}: $\beta_{\mathrm{peak}}\!>\!6^\circ$ is marked
$^{\times}$ (violation), $\beta_{\mathrm{peak}}\!\ge\!15^\circ$ a \emph{spin}. The path-tracking tests (M1,
M2)---the only maneuvers whose ISO~3888 courses define a reference path, the rest prescribing the steering
input directly (Table~\ref{tab:scenarios})---additionally report peak lateral path deviation (devMax [m]),
feeding the summary row of Table~\ref{tab:cmp-7dof}. Scoring the heterogeneous M1--M7 set by
this \emph{one} common metric---rather than each standard's maneuver-specific native measure---makes the
NOM$\to$D-res comparison directly commensurable across maneuvers. Fleet medians carry a five-seed
sampling-stability check ($\pm0.014^\circ$, $\le\!0.15^\circ$ per cell); significance is assessed by
paired Wilcoxon signed-rank tests, Holm-corrected, with Hodges--Lehmann effect sizes.

The ablation isolates the two ingredients of D-res---the learned achievable-yaw reference (the diffusion
\emph{mean}) and the predictive chance back-off (the diffusion \emph{spread})---as a \emph{five}-rung ladder
(Table~\ref{tab:cmp-7dof}) in which every rung shares the \emph{identical} back-off parameters and MPC
weights:
(i)~\textbf{NOM}, the nominal LTV-MPC~\eqref{eq:nominal-mpc} with the linear reference and no back-off;
(ii)~\textbf{D-env}, the predictive chance back-off~\eqref{eq:joint-backoff} on that \emph{same} nominal
reference (spread only)---a diagnostic ablation rung, since the back-off can over-tighten a
reference it cannot re-size, even regressing cells
(Sections~\ref{sub:eval-7dof} and~\ref{sub:cmp-cm});
(iii)~\textbf{D-ref}, the diffusion achievable-yaw reference~\eqref{eq:diffref} \emph{without} the back-off
(mean only);
(iv)~\textbf{GP-ref}, a cautious Gaussian-process mean under the \emph{same} back-off; and
(v)~\textbf{D-res}, the diffusion mean \emph{with} the back-off (both modules).
The four rungs NOM/D-env/D-ref/D-res form a $2\times2$ mean$\times$spread decomposition, and GP-ref differs
from D-res only in the mean's source, so adjacent gaps isolate single effects: D-res$-$D-env is the learned
mean's contribution, D-res$-$D-ref the second-moment back-off's, and D-res$-$GP-ref the diffusion
generator's.

The cautious Gaussian-process baseline (GP-ref)~\cite{hewing2020cautious} replaces the
conditional-diffusion posterior $p_\psi(\Delta r^{\mathrm{ref}}\mid c)$
in~\eqref{eq:diffref-gen} with a Gaussian-process posterior over the \emph{same} reference
residual~\eqref{eq:refresidual}, fit \emph{per vehicle} on the identical command-conditioned pairs
with an automatic-relevance-determination (ARD) squared-exponential GP~\cite{rasmussen2006gp}. At a query
$c=[\vx,\delta]$ the posterior is Gaussian, its predictive mean and variance
\begin{subequations}\label{eq:gp-post}
  \begin{align}
    m(c)   &= k(c,C)\,[K+\sigma_n^2 I]^{-1}\,\mathbf{y},\label{eq:gp-mean}\\
    s^2(c) &= \sigma_n^2 + k(c,c)-k(c,C)\,[K+\sigma_n^2 I]^{-1}\,k(C,c),\label{eq:gp-var}
  \end{align}
\end{subequations}
with training data $(C,\mathbf{y})$, kernel matrix
$K{=}k(C,C)$, and noise variance $\sigma_n^2$. Both moments are tabulated on the $(\vx,\delta)$
grid and substituted \emph{termwise} for the $S$-draw moments~\eqref{eq:refmoments},
$\Delta\bar{r}^{\mathrm{ref}}\!\to\!m(c)$ and $\sigma_r\!\to\!s(c)$, in~\eqref{eq:diffref}
and~\eqref{eq:yaw-joint}. Every other controller quantity---the propagated state-residual term
$s_{3,k}$~\eqref{eq:beta-yaw-proj}, the risk split $k_i=\Phi^{-1}(1-\eta_i)$, the envelope back-off, and all
MPC weights---is left unchanged, so GP-ref and D-res differ \emph{only} in the source of $(m,s)$.
Deployment consumes only these two moments, so the online controller is estimator-agnostic; the comparison
measures an empirical advantage over this cautious baseline, not a uniqueness claim for diffusion.

The diffusion generator (the conditional DDPM of Section~\ref{sub:residual}) is trained \emph{per tier} on
that tier's reference residuals~\eqref{eq:refresidual} by Adam (rate
$2\times10^{-3}$, batch $256$, $300$ epochs), pooling the
four fleet vehicles (Table~\ref{tab:fleet}) over six maneuvers (\{M1,\,M3--M7\}), filtered to
$\vx\!\ge\!5$\,m/s and $|\beta|\!<\!20^\circ$, balanced and steering-mirrored. Its two
moments~\eqref{eq:refmoments} are sampled offline ($S{=}64$ draws) onto a $31\times37$ $(\vx,\delta)$ grid,
odd-symmetrized so $\Delta\bar{r}^{\mathrm{ref}}(\vx,0){=}0$; we deploy the in-sample fleet generator
and keep the leave-one-vehicle-out (LOVO) variant only as a transfer diagnostic:
held-out reference-residual RMS is largely retained ($\approx\!78\%$) on the three lighter vehicles but
collapses on the heavy SUV, yet closed-loop stability transfers (7-DOF LOVO fleet-median $1.02^\circ$ vs.\
$1.11^\circ$ in-sample, no spins), the inflated held-out spread rendering the
back-off~conservative-but-safe.

All controllers share the same design constants across the fleet: an $N_p{=}18$-step prediction and
$N_c{=}8$-step control horizon at
$T_s{=}10$\,ms, the tabulated moments~\eqref{eq:refmoments}, and the envelope~\eqref{eq:env-facets} fixed
by $\beta_{\max}{=}6^\circ$, $\alpha_{r,\max}{=}4^\circ$ (floor $\alpha_{r,\min}{=}1^\circ$). The
chance back-off~\eqref{eq:joint-backoff} applies one common one-sided level
$\kappa{=}\Phi^{-1}(0.95){=}1.645$ ($\eta{=}0.05$) to all six half-planes, the yaw-bound state
term down-weighted ($k_1{=}0.25\,k_2$), with a shared soft slack $\xi$.

\subsection{Evaluation based on 7-DOF model}\label{sub:eval-7dof}
We evaluate D-res first on a 7-DOF vehicle model in closed loop---a three-DOF rigid body (longitudinal,
lateral, yaw) with four wheel-spin states and combined-slip Magic-Formula
tires~\cite{rajamani2012vehicle,pacejka2012tire}---a license-free tier whose results are reproducible
without commercial simulation tooling.
The ladder in Table~\ref{tab:cmp-7dof} reads monotonically at the fleet level, and \textbf{D-res posts the
lowest fleet-median $\beta_{\mathrm{peak}}$ on every vehicle} ($1.02$--$1.14^\circ$ against NOM's
$1.78$--$3.18^\circ$). The reduction is statistically
significant on both tiers ($p\!<\!10^{-6}$; protocol in Section~\ref{sub:setup}), a typical
vehicle--maneuver case gaining nearly one degree of $\beta_{\mathrm{peak}}$ ($0.87^\circ$). The claim is a
fleet-median one: per cell, D-res is the (co-)lowest rung in $22$ of $40$ and beaten in $18$, chiefly by
GP-ref on the low-$\mu$ steps and the dry brake-in-turn ($26$--$14$ head-to-head). The D-ref$\to$D-res gain
is far smaller (median $0.09^\circ$) yet consistent across cells ($p\!<\!10^{-3}$)---the expected signature
of a back-off that adds margin rather than re-sizing the target (tail compression below).

The two modules contribute unequally and separate cleanly. The chance back-off \emph{alone} (D-env)
cuts the per-vehicle \emph{median} $\beta_{\mathrm{peak}}$ by $18$--$55\%$ over NOM, but is not self-sufficient:
applied to the nominal reference it can \emph{over-tighten} a target it cannot re-size, even
regressing the sedan low-$\mu$ lane change ($2.7\!\to\!4.9^\circ$, M1-low). The learned \emph{mean} alone
(D-ref) instead re-sizes that target directly, cutting the median by $24$--$64\%$ without regressions of that magnitude
and carrying the larger share of the reduction on three of the four vehicles. Combining the two, D-res
reaches $36$--$66\%$. The \emph{marginal} value of the second-moment back-off---the D-res$-$D-ref gap---is
therefore vehicle-dependent: largest on the sedan and compact ($1.37\!\to\!1.02$ and $1.59\!\to\!1.14^\circ$
median), where the back-off contributes envelope safety the mean does not, and smallest on the sports car
($1.13\!\to\!1.09^\circ$), where the mean re-size alone is already near-optimal. In the tail (fleet-max
row), the back-off compresses D-ref's worst case from $3.07$--$3.60^\circ$ to $2.59$--$2.88^\circ$.

Substituting a Gaussian mean
for the diffusion mean under the \emph{same} back-off (GP-ref) adds \emph{less} than D-res on every vehicle;
the gap is starkest on the compact car, where GP-ref gains almost nothing over D-env ($19\%$ vs.\
$18\%$)---its marginal-likelihood-regularized posterior reverting to the prior---while D-res reaches $45\%$.
Re-fitting the GP on the same fleet-pooled pairs the diffusion model sees recovers about half of this
compact-car deficit ($19\%\!\to\!34\%$)---the per-vehicle GP was indeed somewhat data-starved---yet D-res
($45\%$) still leads, so the advantage is genuine rather than a pooling confound.
The GP-ref$\leftrightarrow$D-res gap thus isolates the value of the diffusion generator over a calibrated
Gaussian estimator, largest exactly where the achievable-yaw reference must be re-sized most. There the
near-limit residual departs from a bell shape in exactly the way tire saturation dictates: its errors tail
off toward less yaw than the linear reference promises, and at low $\mu$ its mass piles up against the grip
limit rather than spreading into tails (skewness $-1.34$, excess kurtosis $-1.81$, normality rejected at
$p\!<\!0.001$, distribution not shown). This saturation shape a symmetric Gaussian posterior cannot
represent and instead over-smooths.

Two properties of D-res underlie this result. The chance back-off is \emph{self-gating}: it binds only where
the yaw bound~\eqref{eq:nmpc-yaw} or rear-slip facet~\eqref{eq:nmpc-rslip} is active, adding caution on the
limit and low-$\mu$ maneuvers and receding to the nominal reference on benign ones---and it eliminates the sole
7-DOF envelope exceedance (sports-car M7, $6.07^\circ\!>\!6^\circ$ under NOM), the rear-slip facet's
$\vy$--$r$ cross term---which a yaw-only back-off discards---carrying this brake-in-turn
(Fig.~\ref{fig:dres-3d}). The two learned moments then split by regime: the mean governs the severe lane
change (M2), the back-off the dry transient limit (four-vehicle median M5 $3.66\!\to\!0.93$, M7
$2.91\!\to\!1.81$).

Beyond peak side-slip, the back-off is empirically \emph{calibrated}: across the
scored maneuvers on the design sedan the realized state exits the rear-slip facet for $4.9\%$ of the
samples ($\vx\!>\!5$\,m/s) under NOM---up to $27\%$ on the brake-in-turn---but for $0\%$ under D-res,
against the nominal one-sided risk $\eta{=}5\%$; the deliberately conservative fully-correlated-horizon
covariance~\eqref{eq:proj-cov-Sigma} thus keeps the realized risk within the promised level---audited, not
merely asserted. The
per-step Gaussian tightening is nonetheless optimistic: because the residual is skewed, the margin that
actually covers $95\%$ of the draws is $1.75\hat\sigma$ at the fleet median (up to $3.1\hat\sigma$ near the
limit), not the Gaussian $1.645\hat\sigma$, so the deployed back-off attains its nominal coverage on only
about a quarter of cells. The closed-loop exit rate still meets $\eta$ because the mean re-size governs the
outcome and the back-off is a secondary margin; storing the \emph{measured} $95\%$ quantile (the empirical
VaR) in place of $1.645\hat\sigma$ would restore per-step coverage---an upgrade the
architecture makes free, since the same $S{=}64$ offline draws that fill the $\hat\sigma$ table can fill a
one-sided-quantile table at identical online cost, exactly the non-Gaussian capability a parametric
Gaussian estimator lacks. We retain the Gaussian form here for strict comparability with GP-ref under an
identical back-off.

A fixed
worst-case margin does not substitute for this learned spread. In a dedicated constant-$\sigma$ study on the
dry sedan limit maneuvers, replacing the operating-point-conditioned back-off with a constant $\sigma$ sized
to the fleet residual raises the median $\beta_{\mathrm{peak}}$ to $3.03^\circ$, \emph{worse} than the
$2.48^\circ$ of NOM on that set, as it over-tightens the benign maneuvers without re-sizing the reference;
the conditioned D-res instead reaches $0.83^\circ$, so the gain is the conditioning, not the conservatism. The entire comparison, however, rests on a 7-DOF model---a reduced description of the
real vehicle.
How far the fixed bicycle model our controller predicts with diverges from high-fidelity CarMaker is
quantified in Fig.~\ref{fig:diversity}: the
mismatch is small in the linear regime but grows abruptly toward the limit, where the bicycle over-predicts
the yaw rate by more than $2\times$. The 7-DOF plant narrows this gap but does
not close it, so we turn next to a high-fidelity, in-domain CarMaker evaluation.

\subsection{Evaluation based on CarMaker}\label{sub:cmp-cm}
We evaluate D-res (= D-env + D-ref) on the in-domain CarMaker~13 tier---a high-fidelity multibody
co-simulation (Magic-Formula tires) and the generator's training domain, where tire saturation makes both
the model mismatch and the correction largest. It is the high-fidelity step short of hardware;
hardware-in-the-loop is future work (Section~\ref{sec:conclusion}). Setup, KPI, training, and the
five-rung ladder follow the shared protocol of Section~\ref{sub:setup}; all runs are live
CarMaker-for-Simulink (CM4SL) co-simulations on CarMaker~13 (Table~\ref{tab:cmp-cm}).

The nominal LTV-MPC~\eqref{eq:nominal-mpc} keeps the fleet inside the $\pm6^\circ$ envelope on dry
surfaces, but commanding against $\mu{=}1$ it cannot observe a low-$\mu$ surface and diverges: it reaches
$99$--$125^\circ$ on the low-$\mu$ step (three of four vehicles) and $7.9^\circ$ on the low-$\mu$ DLC
(Table~\ref{tab:cmp-cm}). This is the structural error the learned residual targets.

\begin{figure}[t]
  \centering
  \IfFileExists{fig_traj_m3steplow.png}{%
  \includegraphics[width=0.68\columnwidth]{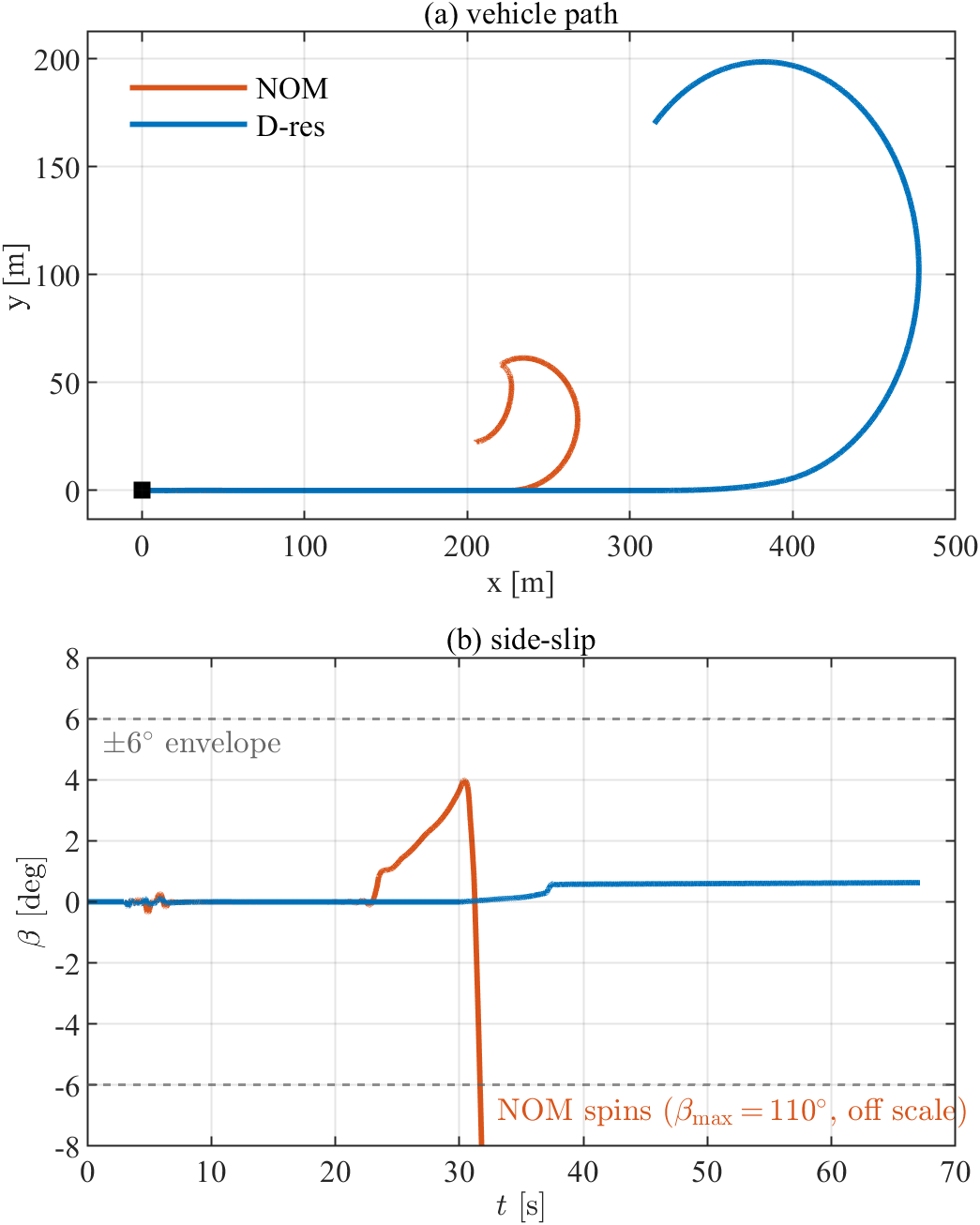}}{%
    \fbox{\parbox[c][0.9\columnwidth][c]{0.98\columnwidth}{\centering\itshape
  [fig\_traj\_m3steplow.png pending --- live CarMaker NOM vs D-res on the SUV M3 step, low-$\mu$]}}}
  \caption{Low-$\mu$ step (M3-low, $\mu\!\approx\!0.3$, SUV), live CarMaker. \textbf{(a)}~Vehicle path:
    NOM vs.\ D-res. \textbf{(b)}~Side-slip $\beta(t)$ (zoomed to
  $\pm8^\circ$).}
  \label{fig:traj-m3low}
\end{figure}

As on the 7-DOF tier the ladder reads monotonically but with a wider gap: \textbf{D-res cuts the
fleet-median $\beta_{\mathrm{peak}}$ by $72$--$84\%$}, substantially exceeding the $36$--$66\%$ of the 7-DOF tier
(Table~\ref{tab:cmp-7dof}), \textbf{and recovers the low-$\mu$ divergence outright}. The envelope back-off
alone (D-env) cuts $6$--$64\%$ over NOM (as little as $6\%$ on the sports car), yet---applied to the
nominal reference---it cannot re-size the friction-limited target and even \emph{regresses} several cells (the sedan low-$\mu$ DLC
$7.9\!\to\!14.5^\circ$, severe DLC $0.8\!\to\!3.5^\circ$). Here the learned \emph{mean} carries the
correction: the diffusion reference \emph{alone} (D-ref) cuts the median by $47$--$81\%$ and turns the
low-$\mu$ step from $99$--$125^\circ$ to $\le\!0.63^\circ$ fleet-wide with no back-off at all---re-sizing
the target to the achievable low-$\mu$ yaw is exactly what that divergence needs
(Section~\ref{sec:method}). Adding the second-moment back-off completes D-res; its \emph{marginal}
contribution (the D-res$-$D-ref gap) concentrates on the dry transient limit and the low-$\mu$ DLC (sedan
M1-low $1.60\!\to\!0.25^\circ$, M5 $1.17\!\to\!0.57^\circ$), turning the low-$\mu$ DLC to $\le\!0.3^\circ$
and holding the dry transient limit (M5,\,M7) near $0.5^\circ$. Figure~\ref{fig:traj-m3low} makes this
concrete on the SUV low-$\mu$ step, where NOM spins ($110^\circ$) while D-res holds the vehicle inside the
envelope ($0.63^\circ$)---the learned reference as the difference between a spin and a lane-keeping recovery.

\emph{Is the diffusion generator necessary, or would a calibrated Gaussian estimator do?} The GP-ref
baseline of Section~\ref{sub:setup} answers this \emph{fleet-wide} here, as on the 7-DOF tier: it replaces
only the uncertainty oracle, so the
GP-ref$\leftrightarrow$D-res gap isolates the diffusion generator on this tier. At the sedan fleet median the
two are close (GP-ref $0.62^\circ$, D-res $0.57^\circ$): a marginal-likelihood-regularized Gaussian is a strong baseline
across the maneuver bulk. But the gap opens---and is wider than on the 7-DOF tier---exactly where the
achievable-yaw reference must be re-sized most aggressively. On the low-$\mu$ DLC (M1-low) the diffusion
reference holds the sedan at $0.25^\circ$ while the smoother GP posterior under-sizes the target and lets
$\beta_{\mathrm{peak}}$ reach $1.60^\circ$ ($6.4\times$ worse); on the sustained-limit circular (M4) the diffusion holds
$0.78^\circ$ against the GP's $1.62^\circ$. The GP betters D-res by a clear margin only on the severe DLC
(M2, $0.58^\circ$ vs.\ $1.55^\circ$); across the remaining sedan cells---away from the re-sizing-sensitive
M1-low and M4---the two track within about a quarter-degree ($\le\!0.25^\circ$).

Beyond the sedan the separation is starker still: on the SUV the GP posterior
\emph{spins} the low-$\mu$ step ($113^\circ$) where the diffusion reference recovers ($0.63^\circ$), and on the
\emph{sports} car the GP reference is so over-smoothed that it fails to improve on the back-off at all---its
fleet median ($2.85^\circ$) is worse than NOM ($2.23^\circ$)---while the diffusion reference reaches
$0.35^\circ$ ($84\%$). Even on the compact car GP-ref ($39\%$) fails to match D-env alone ($43\%$), leaving the
back-off to do the work, whereas D-res reaches $75\%$. The CarMaker head-to-head thus sharpens
the 7-DOF verdict: the diffusion advantage over a Gaussian estimator concentrates in the
aggressive, near-limit regime. The tier contrast is explicit: the diffusion edge over the Gaussian
estimator is only weakly significant on the 7-DOF tier ($p{=}0.035$, typical effect $0.15^\circ$) but
strong on high-fidelity CarMaker ($p\!<\!10^{-5}$, typical effect $0.97^\circ$; $31$--$5$ head-to-head),
where the achievable-yaw map saturates most
sharply and a marginal-likelihood-regularized Gaussian posterior over-smooths its mean back toward the
prior. We accordingly scope the diffusion-over-Gaussian claim to the high-fidelity tier, reading the 7-DOF
gap as suggestive.

On the safety KPIs the mean is already sufficient---D-ref alone posts zero
spins and zero violations here---so the back-off is not what makes the controller safe; its value is margin
and tail, compressing the per-vehicle worst case (fleet-max row) from D-ref's $2.95$/$1.47^\circ$
(compact/sports) to $1.33$/$0.67^\circ$ and carrying the aggregate-audited calibration of
Section~\ref{sub:eval-7dof}. The back-off remains \emph{self-gating} (Section~\ref{sub:eval-7dof}), binding
only on the active friction-limited yaw bound.

Of the envelope facets only the yaw bound is load-bearing: the static side-slip
box~\eqref{eq:nmpc-beta} is redundant---removing it leaves $\beta_{\mathrm{peak}}$ unchanged, the Beal--Gerdes
envelope~\cite{bealgerdes2013mpc} acting through the rear-slip line and the yaw bound alone. The rear-slip
line in turn binds on so few cells that the joint back-off~\eqref{eq:joint-backoff} reduces to its scalar
yaw-only special case ($<\!0.05^\circ$ across the CarMaker sweep), so D-env is effectively the scalar
$s_{3,k}\!=\!\sqrt{\Sigma_{r,k}}$ yaw-bound tightening (driven by $\sigma_{H_r}$) rather than the rear-slip extension. D-res keeps all $36$ envelope-relevant
cells inside the $6^\circ$ envelope, the sole exception being the maximum-amplitude low-$\mu$ sweep
(M6-low), discussed~below.

D-res must also not sacrifice lane-following on the two path-tracking maneuvers (M1, M2). Path deviation is a
joint property of the controller \emph{and} the driver (Section~\ref{sec:problem}): the limited-gain Stanley
follower (7-DOF) cannot fully re-plan around the reduced target, so the back-off trades a modest path margin
for stability (Section~\ref{sub:dref}), whereas the stronger IPGDriver (CarMaker) absorbs the same re-sized
target and keeps D-res's lane deviation at the NOM level. The two drivers differ in the \emph{sign} of the
path-following effect---a property of the driver model, not the controller---yet agree that D-res attains its
side-slip objective within a small, bounded path-following budget.

Within the steering-only scope, the one residual failure is the maximum-amplitude low-$\mu$ sweep
(M6-low, the $^{\dagger}$ cells of Table~\ref{tab:cmp-cm}): its dry-calibrated ISO~7401 amplitude demands a
near-limit lateral acceleration the $\mu\!\approx\!0.3$ surface cannot supply, pinning the cell at the
friction boundary, where outcomes bifurcate: on the sedan and SUV \emph{every} rung spins
($78$--$148^\circ$, NOM and D-res alike), the compact stays near the boundary under every rung
($6.1$--$8.5^\circ$), and on the sports car only D-res returns inside ($2.4^\circ$)---boundary sensitivity,
not a controllable separation, so no steering reference reliably passes. We therefore read M6-low as a
maneuver-design limit rather than a controller failure, shown for completeness and excluded from the
$36/36$ envelope-pass base.

Across the two tiers, then, the \emph{same} controller---identical design constants, only the generator
retrained per tier---holds the stable-handling envelope under two different
driver models (the fixed-gain Stanley follower vs.\ the knowledge-based IPGDriver) and two different
vehicle dynamics (the 7-DOF model vs.\ the CarMaker multibody). The remaining perturbation axis---tire and
load on a fixed vehicle---is probed next.

\subsection{Robustness to load and tire uncertainty}\label{sub:abl-diversity}
Section~\ref{sec:problem} identified the vehicle axis---class, tire, and load shifting the effective
stiffnesses---as a primary source of the model uncertainty the learned correction must absorb. We finally
assess the robustness of D-res to the tire and load dimensions of that axis, whose class dimension is
resolved separately by the per-vehicle columns of Tables~\ref{tab:cmp-7dof} and~\ref{tab:cmp-cm}.
We test whether the command-conditioned
correction, trained across the fleet, holds the envelope as these perturbations shift the effective
stiffnesses \emph{without re-identifying the model} (Fig.~\ref{fig:diversity}). On the design sedan we vary the tire (low-friction
Tire~B, 195/65R15) and the load ($+280$\,kg, five passengers) on the limit-stressing maneuvers M1, M5, and
M7 under NOM and D-res (Table~\ref{tab:abl-diversity}).

Against the Tire-A, one-passenger baseline, the low-friction tire and the added load each raise the
NOM peak side-slip on every maneuver---to $3.55^\circ$ (Tire~B) and $3.26^\circ$ (load)---and
stacking both drives NOM to $4.07^\circ$ on the brake-in-turn M7, so plain feedback does not absorb the
perturbation. D-res restores the envelope throughout, holding $\beta_{\mathrm{peak}}\!\le\!1.35^\circ$ on
\emph{every} cell, including the compound Tire~B\,$+$\,load condition on which it was never explicitly
trained. The residual conditioned only on the command thus absorbs the tire- and load-induced stiffness
shift of Fig.~\ref{fig:diversity}, providing robustness to \emph{model} uncertainty in place of
re-identification or a worst-case bound.

\begin{table}[t]
  \centering
  \caption{Load/mass/tire diversity on the design sedan (CarMaker~13): peak side-slip $\beta_{\mathrm{peak}}$
  [deg]; the baseline row repeats the design-sedan entries of Table~\ref{tab:cmp-cm}.}
  \label{tab:abl-diversity}
  \renewcommand{\arraystretch}{1.2}
  \setlength{\tabcolsep}{3pt}
  \scriptsize
  \begin{tabular}{@{}l cc cc cc@{}}
    \toprule
    & \multicolumn{2}{c}{M1} & \multicolumn{2}{c}{M5} & \multicolumn{2}{c}{M7} \\
    \cmidrule(lr){2-3}\cmidrule(lr){4-5}\cmidrule(lr){6-7}
    Condition & NOM & D-res & NOM & D-res & NOM & D-res \\
    \midrule
    Baseline (Tire~A, 1-pax)    & $1.54$ & $\mathbf{0.72}$ & $2.64$ & $\mathbf{0.57}$ & $2.95$ & $\mathbf{0.43}$ \\
    Tire~B (low-$\mu$)          & $2.92$ & $\mathbf{1.09}$ & $3.55$ & $\mathbf{0.92}$ & $3.30$ & $\mathbf{0.64}$ \\
    Load ($+280$\,kg)           & $2.01$ & $\mathbf{0.84}$ & $3.26$ & $\mathbf{0.73}$ & $3.11$ & $\mathbf{0.51}$ \\
    Tire~B\,$+$\,load (worst)   & $3.05$ & $\mathbf{1.35}$ & $3.38$ & $\mathbf{1.17}$ & $4.07$ & $\mathbf{0.81}$ \\
    \bottomrule
  \end{tabular}
\end{table}

\subsection{On-target real-time deployment}\label{sub:realtime}
The offline tabulation makes the online controller cheap enough for an embedded automotive processor. We
deploy the \emph{full} D-res step---the LTV-MPC QP, the chance back-off, and the $O(1)$ moment lookup, with
no in-loop diffusion---to an NVIDIA Jetson AGX Xavier, generated from the MATLAB source with MATLAB Coder
(R2024b) as portable single-core scalar C (ARM Cortex-A, gcc~\texttt{-O3}) on one Carmel core at
$2.27$\,GHz; the tabulated moments occupy about $37$\,kB. The step is deterministic: a small condensed QP
(eight decision variables, $133$ dual constraints) solved by a \emph{fixed-iteration} Hildreth method makes
the per-step time input-independent, so a single trace characterizes the fleet and both tiers, and the
generated C matches the reference \texttt{quadprog} solution to $0.000^\circ$---the on-target controller is
the \emph{same} one that produces Tables~\ref{tab:cmp-7dof} and~\ref{tab:cmp-cm}.

Timing is measured with \texttt{CLOCK\_MONOTONIC} over $12{,}000$ steps of a live trace on the
sine-with-dwell (M5) and brake-in-turn (M7) maneuvers, whose constraint-active, tire-saturated segments
exercise the solver's worst case. At the accuracy-validated setting (Table~\ref{tab:wcet}) the controller
completes in a worst-case $\mathbf{4.08}$\,\textbf{ms} ($41\%$ of the budget) with a $3.1$\,ms mean; fewer
iterations do not converge at the limit and more only erode the margin. This is on-target \emph{profiling},
not a closed-loop hardware-in-the-loop test, and the WCET is a measured maximum rather than a statically
certified bound---both left to future work; even so, the controller meets the $100$\,Hz requirement on a
single core, in unoptimized scalar C, with more than half the budget to spare.

\begin{table}[t]
  \centering
  \caption{On-target per-step execution time of the deployed D-res controller
    versus the fixed Hildreth iteration count. WCET is the observed maximum
    over $12{,}000$ steps; the accuracy column marks agreement with the \texttt{quadprog} reference at the
  limit of handling.}
  \label{tab:wcet}
  \renewcommand{\arraystretch}{1.2}
  \setlength{\tabcolsep}{6pt}
  \footnotesize
  \begin{tabular}{@{}c c c c c@{}}
    \toprule
    iter & WCET [ms] & \% of $10$\,ms & mean [ms] & accuracy \\
    \midrule
    $40$          & $2.33$          & $23\%$          & $1.61$          & $\times$~($0.69^\circ$) \\
    $\mathbf{80}$ & $\mathbf{4.08}$ & $\mathbf{41\%}$ & $\mathbf{3.11}$ & $\checkmark$~(deployed) \\
    $100$         & $5.37$          & $54\%$          & $3.85$          & $\checkmark$ \\
    $160$         & $8.10$          & $81\%$          & $6.08$          & $\checkmark$ \\
    \bottomrule
  \end{tabular}
\end{table}

%% file: sections/07-conclusion.tex
\section{Conclusion}\label{sec:conclusion}

We have presented diffusion-residual model predictive steering control (D-res) for vehicle stabilization
at the limit of handling. A conditional diffusion residual model, conditioned on the steering command,
provides both the yaw-rate reference through its mean and a one-sided chance back-off on the
stable-handling envelope through its calibrated spread, entering the controller's reference and constraints
rather than its control law. Evaluated across vehicle, tire,
road, and maneuver diversity on a 7-DOF model and high-fidelity CarMaker co-simulation, D-res reduces peak
side-slip where the fixed bicycle model is least accurate and restores directional stability on
low-friction maneuvers on which the fixed reference over-commands the available grip. Because the online
controller adds only a single table lookup to the baseline MPC, it runs within the $100$\,Hz budget on an
NVIDIA Jetson AGX Xavier at a measured worst-case $4.08$\,ms per step, with no in-loop~diffusion.

Two limitations bound the steering-only scope adopted here. The maximum-amplitude, low-friction
frequency sweep (M6-low) is a maneuver-design limit at the friction boundary rather than a controller
failure; and, beyond the evaluated matrix, a low-friction brake-in-turn would pose a longitudinal and
load-transfer problem beyond the authority of front steering. Future
work will extend the learned residual to an integrated-chassis setting that couples steering with
longitudinal control and differential braking, and will incorporate a learned road-friction estimate to
bound the yaw rate where longitudinal authority can act upon it. We further aim---having demonstrated
on-target real-time execution (Section~\ref{sub:realtime})---to close the loop with hardware-in-the-loop
and vehicle testing and a static worst-case-timing certification.